\documentclass[11pt, letterpaper]{cmu}
\pdfoutput=1
\usepackage{etoolbox,verbatim}
\usepackage{wrapfig}
\usepackage{needspace}

\newtoggle{arxiv}
\toggletrue{arxiv}

\newtoggle{customthms}
\toggletrue{customthms}

\newtoggle{colt}
\togglefalse{colt}

\usepackage[numbers,sort&compress]{natbib}

\usepackage{xcolor}
\usepackage{tcolorbox}
\tcbuselibrary{skins,breakable}

\definecolor{rliableolive}{HTML}{BBCC33}
\tcbset{
  aibox/.style={
      width=\linewidth,
      top=3pt,
      bottom=2pt,
      before skip=6pt plus 2pt,
      after skip=8pt plus 2pt,
      colback=rliableolive!8!white,
      colframe=black,
      colbacktitle=black,
      enhanced,
      center,
      attach boxed title to top left={yshift=-0.1in,xshift=0.15in},
      boxed title style={boxrule=0pt,colframe=white,},
    }
}
\newtcolorbox{AIbox}[2][]{aibox,title=#2,#1}

\definecolor{primalcolor}{HTML}{A60000}
\definecolor{contrarycolor}{HTML}{00A6A6}
\definecolor{darkcontrarycolor}{HTML}{004C4C}
\definecolor{lightblue}{HTML}{2970CC}
\definecolor{lightpurple}{HTML}{673147}
\definecolor{ForestGreen}{HTML}{FF5733}
\definecolor{myred}{HTML}{AA4A44}
\definecolor{hyppurple}{HTML}{800080}

\newcommand{\thmcolordark}{red!30!black}

\usepackage[T1]{fontenc}
\usepackage{capt-of}
\usepackage{scalefnt}
\usepackage{amssymb,upgreek,bm}
\usepackage{mathtools}
\usepackage[english]{babel}  
\usepackage{multirow,tcolorbox,tikz-cd,turnstile,xspace,xparse}
\usepackage{relsize,breakcites,appendix}
\usepackage{longtable,tablefootnote,booktabs,float,makecell}
\usepackage[normalem]{ulem}
\usepackage{tabu}
\usepackage{enumitem}
\usepackage{footnote}
\usepackage{boxedminipage}
\usepackage{multirow,nicefrac}
\usepackage{verbatim} 

\hypersetup{
    colorlinks = true,
    citecolor = {magenta},
    linkcolor = {blue},
    urlcolor = {blue},
    breaklinks = true,
}

\usepackage[capitalise,nameinlink]{cleveref}
\usepackage{prettyref}
\iftoggle{colt}{
    \DeclareRobustCommand{\qed}{
        \usepackage{thmtools}
          \ifmmode \mathqed
          \else
            \leavevmode\unskip\penalty9999 \hbox{}\nobreak\hfill
            \quad\hbox{\qedsymbol}%
          \fi
    }
}
{
    \usepackage{amsthm}
     
}
\usepackage{thm-restate}

\iftoggle{arxiv}
{
\usepackage{mathrsfs}

    \usepackage{subcaption}

    \usepackage[noend]{algpseudocode}
    \usepackage{algorithm}
   
}
{
    \usepackage{mathrsfs}
    \usepackage{subcaption}
}

\DeclareMathAlphabet{\mathbfsf}{\encodingdefault}{\sfdefault}{bx}{n}

\numberwithin{equation}{section}

\iftoggle{arxiv}
{
    }
{
    
}
\iftoggle{arxiv}
{
    \newcommand{\colorpar}[1]{
    \paragraph{#1}}
}
{
    \newcommand{\colorpar}[1]{\paragraph{#1}}
}

\usepackage{thmtools}

\Crefname{equation}{Eq.}{Eqs.}
\Crefname{assumption}{Assumption}{Assumptions}
\Crefname{condition}{Condition}{Conditions}
\Crefname{claim}{Claim}{Claims}
\Crefname{property}{Property}{Properties}
\Crefname{construction}{Construction}{Constructions}

\declaretheoremstyle[
    headformat=\normalfont\textcolor{\thmcolordark}{\bfseries\NAME\,\NUMBER}\NOTE,%
    notefont={\normalfont\textcolor{\thmcolordark}{\bfseries}}, 
    notebraces={}{},
    bodyfont=\normalfont\itshape,
    spaceabove = 6pt,
    spacebelow = 6pt,
    ]{coloredthmversion}

\declaretheoremstyle[
    headformat=\normalfont\textcolor{\thmcolordark}{\bfseries\NAME\,\NUMBER}\NOTE,%
    bodyfont=\normalfont\itshape,
    spaceabove = 6pt,
    spacebelow = 6pt,
    ]{coloredthm}

\declaretheoremstyle[
    headformat=\normalfont\textcolor{\thmcolordark}{\bfseries\NAME\,\NUMBER}\NOTE,%
    bodyfont=\normalfont,
    spaceabove = 6pt,
    spacebelow = 6pt,
    ]{coloreddef}

\iftoggle{customthms}
{
    \theoremstyle{coloredthmversion}
}
{}

\iftoggle{customthms}{
  \theoremstyle{coloredthm}
  \newtheorem{theorem}{Theorem}
  \newtheorem{lemma}{Lemma}[section]
  \newtheorem{corollary}{Corollary}[section]
  \newtheorem{proposition}[lemma]{Proposition}
}
{}

\iftoggle{customthms}{
    \theoremstyle{coloreddef}
    \newtheorem{definition}{Definition}[section]

    \newtheorem{property}{Property}[section]
}
{
  
}

\newtheorem{assumption}{Assumption}[section]
\newtheorem{condition}{Condition}[section]

\makeatletter
\newcommand{\neutralize}[1]{\expandafter\let\csname c@#1\endcsname\count@}
\makeatother

\iftoggle{customthms}{
    \newtheoremstyle{named}{}{}{\itshape}{}{\bfseries}{}{.5em}{\Cref{#3} {\normalfont (informal)} }{}
    \theoremstyle{named}
    
    \theoremstyle{plain}
}
{}

\newtheorem*{theorem*}{Theorem}
\newtheorem*{lemma*}{Lemma}
\newtheorem*{corollary*}{Corollary}
\newtheorem*{proposition*}{Proposition}
\newtheorem*{claim*}{Claim}
\newtheorem*{fact*}{Fact}
\newtheorem*{observation*}{Observation}
\newtheorem*{definition*}{Definition}
\newtheorem*{remark*}{Remark}
\newtheorem*{example*}{Example}

\def\ddefloop#1{\ifx\ddefloop#1\else\ddef{#1}\expandafter\ddefloop\fi}
\def\ddef#1{\expandafter\def\csname bb#1\endcsname{\ensuremath{\mathbb{#1}}}}
\ddefloop ABCDEFGHIJKLMNOPQRSTUVWXYZ\ddefloop

\def\ddefloop#1{\ifx\ddefloop#1\else\ddef{#1}\expandafter\ddefloop\fi}
\def\ddef#1{\expandafter\def\csname frak#1\endcsname{\ensuremath{\mathfrak{#1}}}}
\ddefloop ABCDEFGHIJKLMNOPQRSTUVWXYZ\ddefloop

\def\ddefloop#1{\ifx\ddefloop#1\else\ddef{#1}\expandafter\ddefloop\fi}
\def\ddef#1{\expandafter\def\csname fr#1\endcsname{\ensuremath{\mathfrak{#1}}}}
\ddefloop ABCDEFGHIJKLMNOPQRSTUVWXYZ\ddefloop

\def\ddefloop#1{\ifx\ddefloop#1\else\ddef{#1}\expandafter\ddefloop\fi}
\def\ddef#1{\expandafter\def\csname eul#1\endcsname{\ensuremath{\EuScript{#1}}}}
\ddefloop ABCDEFGHIJKLMNOPQRSTUVWXYZ\ddefloop

\def\ddefloop#1{\ifx\ddefloop#1\else\ddef{#1}\expandafter\ddefloop\fi}
\def\ddef#1{\expandafter\def\csname scr#1\endcsname{\ensuremath{\mathscr{#1}}}}
\ddefloop ABCDEFGHIJKLMNOPQRSTUVWXYZ\ddefloop

\def\ddefloop#1{\ifx\ddefloop#1\else\ddef{#1}\expandafter\ddefloop\fi}
\def\ddef#1{\expandafter\def\csname b#1\endcsname{\ensuremath{\mathbf{#1}}}}
\ddefloop ABCDEFGHIJKLMNOPQRSTUVWXYZ\ddefloop

\def\ddefloop#1{\ifx\ddefloop#1\else\ddef{#1}\expandafter\ddefloop\fi}
\def\ddef#1{\expandafter\def\csname bhat#1\endcsname{\ensuremath{\hat{\mathbf{#1}}}}}
\ddefloop ABCDEFGHIJKLMNOPQRSTUVWXYZ\ddefloop

\def\ddefloop#1{\ifx\ddefloop#1\else\ddef{#1}\expandafter\ddefloop\fi}
\def\ddef#1{\expandafter\def\csname btil#1\endcsname{\ensuremath{\tilde{\mathbf{#1}}}}}
\ddefloop ABCDEFGHIJKLMNOPQRSTUVWXYZ\ddefloop

\def\ddefloop#1{\ifx\ddefloop#1\else\ddef{#1}\expandafter\ddefloop\fi}
\def\ddef#1{\expandafter\def\csname bst#1\endcsname{\ensuremath{\mathbf{#1}^\star}}}
\ddefloop ABCDEFGHIJKLMNOPQRSTUVWXYZ\ddefloop

\def\ddefloop#1{\ifx\ddefloop#1\else\ddef{#1}\expandafter\ddefloop\fi}
\def\ddef#1{\expandafter\def\csname bst#1\endcsname{\ensuremath{\mathbf{#1}^\star}}}
\ddefloop abcdeghijklmnopqrstuvwxyz\ddefloop

\def\ddefloop#1{\ifx\ddefloop#1\else\ddef{#1}\expandafter\ddefloop\fi}
\def\ddef#1{\expandafter\def\csname bhat#1\endcsname{\ensuremath{\hat{\mathbf{#1}}}}}
\ddefloop abcdefghijklmnopqrstuvwxyz\ddefloop

\def\ddefloop#1{\ifx\ddefloop#1\else\ddef{#1}\expandafter\ddefloop\fi}
\def\ddef#1{\expandafter\def\csname b#1\endcsname{\ensuremath{\mathbf{#1}}}}
\ddefloop abcdeghijklnopqrstuvwxyz\ddefloop

\def\ddefloop#1{\ifx\ddefloop#1\else\ddef{#1}\expandafter\ddefloop\fi}
\def\ddef#1{\expandafter\def\csname barb#1\endcsname{\ensuremath{\bar{\mathbf{#1}}}}}
\ddefloop abcdefghijklmnopqrstuvwxyz\ddefloop

\def\ddef#1{\expandafter\def\csname c#1\endcsname{\ensuremath{\mathcal{#1}}}}
\ddefloop ABCDEFGHIJKLMNOPQRSTUVWXYZ\ddefloop
\def\ddef#1{\expandafter\def\csname h#1\endcsname{\ensuremath{\widehat{#1}}}}
\ddefloop ABCDEFGHIJKLMNOPQRSTUVWXYZ\ddefloop
\def\ddef#1{\expandafter\def\csname hc#1\endcsname{\ensuremath{\widehat{\mathcal{#1}}}}}
\ddefloop ABCDEFGHIJKLMNOPQRSTUVWXYZ\ddefloop
\def\ddef#1{\expandafter\def\csname t#1\endcsname{\ensuremath{\widetilde{#1}}}}
\ddefloop ABCDEFGHIJKLMNOPQRSTUVWXYZ\ddefloop
\def\ddef#1{\expandafter\def\csname tc#1\endcsname{\ensuremath{\widetilde{\mathcal{#1}}}}}
\ddefloop ABCDEFGHIJKLMNOPQRSTUVWXYZ\ddefloop

\iftoggle{arxiv}
{
\newcommand{\confpar}[1]{\paragraph{#1}}
}
{
\newcommand{\confpar}[1]{\textbf{#1}}
}

\usepackage{tcolorbox}
\tcbuselibrary{skins,breakable}

\tcbset{
  aibox/.style={
      width=\linewidth,
      top=6pt,
      bottom=0pt,
      left=1.5pt,
      right=1.5pt,
colback=blue!60!gray!5,
colframe=black,
      colbacktitle=black,
      enhanced,
      center,
      attach boxed title to top left={yshift=-0.1in,xshift=0.15in},
      boxed title style={boxrule=0pt,colframe=white,},
    }
}

\iftoggle{arxiv}
{

}
{

}

\iftoggle{arxiv}
{
\renewcommand{\confpar}[1]{\colorpar{#1}}
}
{
\renewcommand{\confpar}[1]{\colorpar{#1}}
}

\DeclareMathOperator*{\argmax}{arg\,max}

\newcommand{\ballkr}[1][r]{\cB_{k}(r)}

\DeclareMathSymbol{\shortminus}{\mathbin}{AMSa}{"39}

\newcommand{\R}{\mathbb{R}}

\newcommand{\pibase}{\pi_{\mathrm{base}}}

\newcommand{\pifive}{\pi_{\mathrm{0.5}}}

\newcommand{\pires}{\pi_{\theta}^{\mathrm{res}}}
\newcommand{\abase}{\ba^{\mathrm{base}}}
\newcommand{\PSName}{minimal-data adaptation\xspace}
\newcommand{\PSNameCaps}{Minimal Data Adaptation }

\newcommand{\midas}{\textsc{MiDAS}\xspace}
\newcommand{\pms}[1]{\ensuremath{\pm #1}}

\usepackage{preamble/color-edits}

\addauthor{ms}{magenta}
\addauthor{sk}{green}

\addauthor{sv}{blue}

\newcommand{\ignore}[1]{}

\usepackage{parskip}

\title{What Does It Take For A Generalist Robot Policy To Adapt Using Minimal Data?}

\author[*]{Shreyas Kowshik}
\author[*]{Sreyas Venkataraman}
\author[{{}}]{Leo Wang}
\author[{{}}]{Niharika Pant}
\author[$\dagger$]{Max Simchowitz}
\author[$\dagger$]{Aviral Kumar}

\affil[{{}}]{Carnegie Mellon University}
\affil[*]{Equal contribution}
\affil[$\dagger$]{Equal advising}

\correspondingauthor{sreyasv@andrew.cmu.edu, skowshik@andrew.cmu.edu, \\ Additional real-world videos and rollouts are available on the website:
\href{https://minimal-data-adaptation.github.io/}{minimal-data-adaptation.github.io}.}

\begin{document}

\begin{abstract}
\textbf{Abstract.} A central goal in robot learning is to move beyond task-specific human data collection toward robots that improve through autonomous interaction. Yet fully autonomous learning remains difficult with current policies: sparse rewards and weak zero-shot exploration often prevent online learning from discovering successful behavior. As a first step towards fully autonomous learning, we propose the \textbf{minimal-data adaptation} regime, in which a pretrained robot policy must learn a new task from as little as one demonstration followed by autonomous online interaction. 
We establish that, although pretrained policies fail zero-shot on many new tasks, a single demonstration suffices to direct their behavioral prior to coarse task-following behavior, albeit with unreliable performance. From this insight, we isolate the key design principles required to convert coarse task-following into reliable task success. These include (1) the surprising efficacy of pretrained VLA representations to support sample-efficient value learning from online interaction (2) the sufficiency of residual policy learning to learn corrective behavior and (3) the need for careful data-balancing to prevent exploration collapse. As a representative of these principles, we propose
\textbf{\midas}, a simple offline-to-online RL recipe that combines few-demonstration behavior cloning with value-based residual learning over frozen VLA representations. Across LIBERO and RoboCasa tasks, \midas substantially improves performance from single-demonstration initializations, including policies that succeed only rarely, validating the principles highlighted above. On a bimanual YAM platform, starting from a brittle initial policy, \midas improves task success after approximately six hours of online interaction. 
\end{abstract}
\maketitle

\section{Introduction}


A long-standing goal in robotics is to develop robots that improve through their own experience: a robot enters a new environment, attempts a task, and autonomously refines its behavior until it can perform the task reliably. Achieving this goal poses challenges ranging from physical considerations, such as providing resets and managing hardware instrumentation, to algorithmic questions about how learning should be initialized and carried out. In this work, we focus on a central algorithmic challenge: enabling rapid adaptation to a new task, with minimal task-specific supervision or design. Prior work on autonomous robot adaptation has largely relied on specialist, single-task policies~\citep{luo2025serlsoftwaresuitesampleefficient,haldar2023teachrobotfishversatile,vosylius2025instantpolicyincontextimitation}, often coupled with substantial task-specific engineering. The emergence of generalist robot policies pretrained on broad and diverse datasets~\citep{octomodelteam2024octoopensourcegeneralistrobot,brohan2023rt1roboticstransformerrealworld,brohan2023rt2visionlanguageactionmodelstransfer,physicalintelligence2025pi05,black2024pi0,ye2026worldactionmodelszeroshot,kim2025openvlaoft} provides an opportunity to revisit this problem: can the broad prior encoded by a pretrained generalist policy enable efficient online adaptation with \textbf{\emph{only minimal task-specific supervision}}?

\providecommand{\mainfigcaption}{%
    \textbf{Overview of \PSName.}
    A single demonstration is sparse and underdetermined on its own, but
    pretrained VLA representations turn it into a useful anchor for adaptation.
    Few-demo behavior cloning narrows the search space to a task-relevant
    region around the demonstration, and online residual RL then expands
    local success by learning value-guided corrections on top of the
    warm-started base policy.}

\iftoggle{arxiv}{%
    \ifcsdef{mainfigrendered}{}{%
        \begin{figure*}[t]
        \centering
        \vspace{-0.2cm}
        \includegraphics[width=\textwidth]{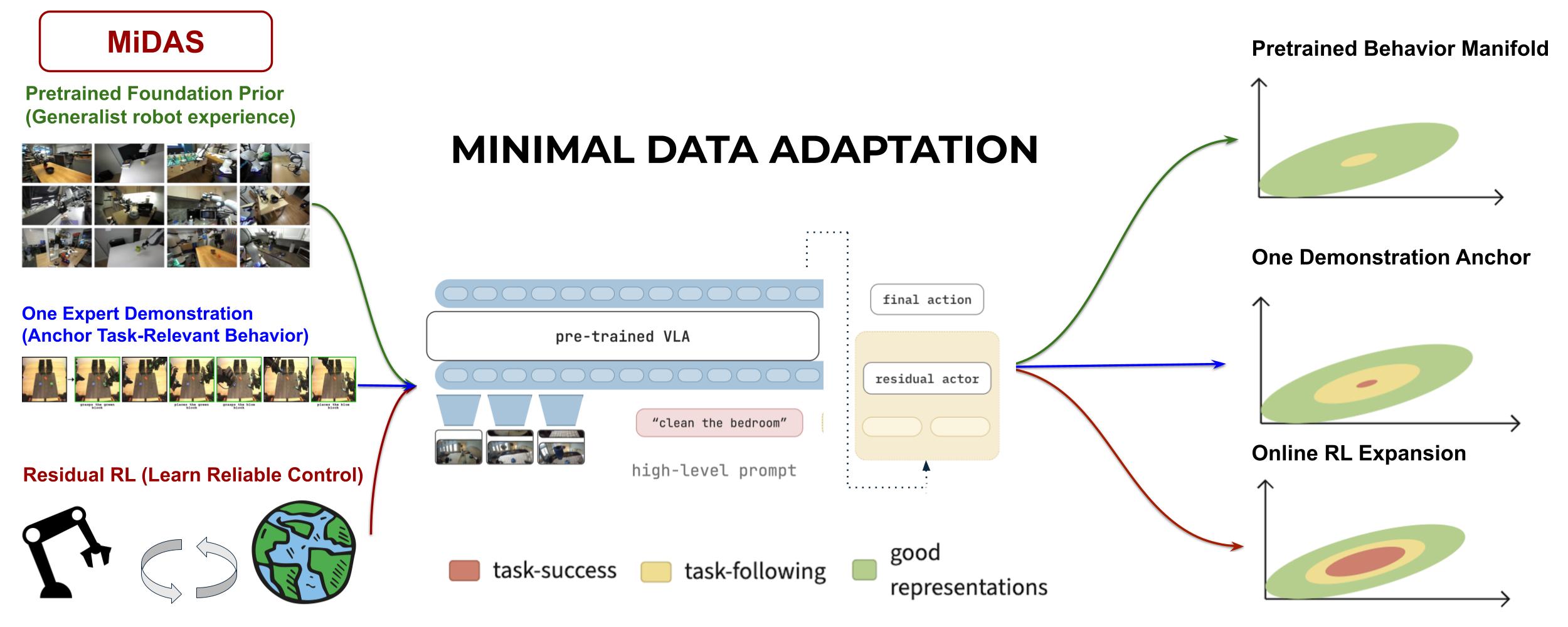}
        \caption{\mainfigcaption}
        \label{fig:main_overview}
        \vspace{-0.1cm}
        \end{figure*}
        \csgdef{mainfigrendered}{}%
    }%
}{%
    \begin{figure*}[ht]
        \vspace{-0.2cm}
        \centering
        \includegraphics[width=\textwidth]{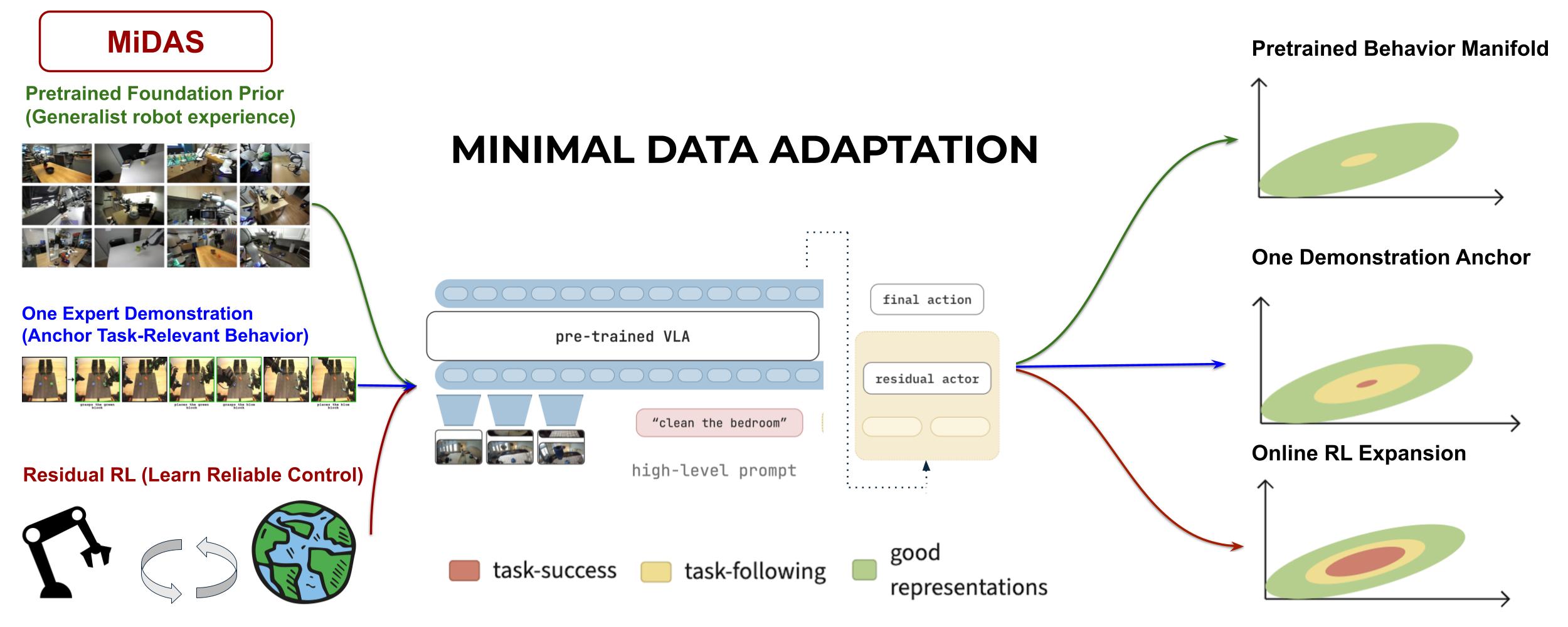}
        \caption{\mainfigcaption}
        \label{fig:main_overview}
        \vspace{-1em}
    \end{figure*}
}

To answer this question, we first note that autonomous adaptation depends on three ingredients. The policy must represent high-dimensional observations in a form useful for policy and critic learning, produce behavior that reaches reward-bearing regions during deployment, and use this experience to improve efficiently.
Given the strong downstream transfer enabled by large-scale pretraining in language, vision, and representation learning \citep{devlin-etal-2019-bert, radford2021learningtransferablevisualmodels, NEURIPS2020_1457c0d6}, one would expect recent generalist robot policies ~\citep{octomodelteam2024octoopensourcegeneralistrobot,brohan2023rt1roboticstransformerrealworld,brohan2023rt2visionlanguageactionmodelstransfer,physicalintelligence2025pi05,black2024pi0,kim2025openvlaoft} to make progress on the first two ingredients: task-relevant representations for control and behavioral (action) priors acquired from diverse manipulation data.
However, these policies are still not reliable zero-shot learners in new environments. They may fail to follow the task instruction, miss sparse reward entirely, or visit only a narrow set of useful states, leaving online learning with too little signal to bootstrap improvement.

We study this gap by operating in the \emph{\textbf{minimal-data adaptation (MDA)}} setting, in which a pretrained policy must learn to improve autonomously from only minimal task-specific supervision, such as a single successful demonstration. Prior work has developed effective methods for autonomous policy improvement, but these methods generally operate in more favorable regimes.
They typically assume either that the initial policy already has substantial task competence~\citep{wagenmaker2025steeringdiffusionpolicylatent,ren2024diffusionpolicypolicyoptimization,hu2024imitationbootstrappedreinforcementlearning, patil2026ogposampleefficientfullfinetuning} or that tens to hundreds of demonstrations are available~\citep{kumar2022ptr, nakamoto2024calqlcalibratedofflinerl,hu2024imitationbootstrappedreinforcementlearning,ball2023efficientonlinereinforcementlearning}. While MDA resembles prior autonomous adaptation settings, minimal task-specific supervision makes stable online learning substantially harder: under sparse rewards, a weak initial policy collects mostly failures, often leading to policy collapse. As a result, methods designed for stronger initial policies or larger demonstration sets do not transfer cleanly to this regime. At the same time, few-shot imitation learning methods suggest that learning from one or two demonstrations is possible in some settings \citep{duan2017oneshotimitationlearning,finn2017oneshotvisualimitationlearning,dipalo2024keypointactiontokensenable}, but most available VLA policies do not naturally exhibit consistent few-shot improvement. MDA targets the narrow regime between zero-shot deployment and standard task-specific finetuning: the pretrained policy is not reliable enough to deploy directly, but may be close enough that one demonstration can make online learning tractable.

Our main finding is that, in many settings, minimal-data adaptation is tractable for today's generalist robot policies. We show that, while behavior cloning of a single  demonstration of a successful trajectory does not provide enough coverage for high rates of task success, a  suitably designed online interaction phase can achieve reliable task performance. In particular, \emph{\textbf{our work isolates how broad pretraining, minimal per-task demonstration, and online interaction jointly enable adaptation.}} We establish that (1) pretraining features provide not just visual information, but task-relevant features that are essential for learning task-specific critics from limited data
(2) that value-based actor-critic policy learning, but not filtered supervised imitation of a policy's own success, affords sufficient exploration to increase task success and (3) how offline and online data should be carefully balanced to prevent being overly constrained by the former, or collapsing by overfitting to the latter.
Together, we consolidate these principles in a representative algorithm we call \textbf{\midas}: an offline-to-online recipe that first fine-tunes a pretrained VLA with imitation learning on $K$ demonstrations to obtain a policy with coarse behaviours, then freezes the VLA and trains a lightweight residual actor on its representations using value-based RL~\citep{mark2024policyagnosticrloffline}.
Moreover, we carefully test the capabilities and limitations of the generalization afforded by adaptation, showing that: (1) robustness to visual and language changes is largely preserved from pretraining; (2) online learning makes the task-relevant behavior induced by the demonstration more reliable across nearby object configurations; (3) gradually expanding the reset distribution during online interaction broadens this generalization without additional demonstrations, but settings that require new manipulation strategies remain challenging.




We make three contributions. \textbf{First}, we introduce \emph{minimal-data adaptation}, a regime where a pretrained robot policy is not reliable zero-shot, but may be close enough that one or a few demonstrations make online learning tractable.
\textbf{Second}, we show that \emph{\textbf{\PSName is possible from as little as 1 demonstration.}} We do so by studying a set of algorithmic choices that enable learning in this regime and identify \midas, a simple value-based RL recipe combining behavior cloning with frozen-backbone residual RL and stabilizing choices for sparse-reward adaptation.
On LIBERO-Long and RoboCasa,
we show that MiDAS recovers substantial task success from policies initialized with a
single demonstration, even when behavior cloning yields an initial success rate below 5\%, significantly outperforming imitation and steering baselines. \midas transfers to a real bimanual YAM platform with roughly 6 hours of autonomous interaction.
\textbf{Third}, we empirically characterize the contributions of demonstrations to task-level behavior, pretrained VLA representations to sample-efficient value learning, and online residual RL to corrective actions beyond those the few-demo policy reliably produces.
We systematically study the capabilities and limitations of generalization achieved by successful adaptation in the MDA regime.

\newcommand{\Ddemo}{\mathcal{D}_{\mathrm{demo}}}
\newcommand{\Mtarget}{\mathcal{M}_{\mathrm{target}}}

\section{Problem Setting: \PSNameCaps}
\label{sec:problem_setting}

We now define the \textbf{minimal-data adaptation} (MDA) setting and motivate why it is a useful regime to study. Given a pretrained generalist robot policy $\pibase$, $K$ successful demonstrations $\Ddemo^K$ collected in a target environment (denoted by $\Mtarget$), and access to autonomous interaction with $\Mtarget$, minimal-data adaptation asks whether we can learn an adapted policy $\pi_\theta^K$ that achieves high return in $\Mtarget$ using only this limited supervision. The goal is not to solve the task from demonstrations alone. Instead, the demonstrations serve as a minimal scaffold that makes subsequent autonomous learning tractable. The case $K{=}1$ is the sharpest version of this setting: learning on a single demonstration is often insufficient to learn previously unseen tasks, but may still expose enough task-relevant behavior for interaction-driven improvement to begin.

MDA targets the regime where autonomous learning first becomes tractable. With no task-specific data, current pretrained policies often fail to reach task-relevant behavior often enough for sparse-reward interaction to improve them. A small number of demonstrations can make the problem just structured enough: they expose the intended behavior and bring the policy into the vicinity of success, while still leaving robustness and distribution coverage to be learned through autonomous interaction. This setting therefore tests whether pretrained robot policies have reached a regime where a few demonstrations can specify the task while online interaction supplies the missing robustness. It also separates the roles of the available sources of supervision: the structure inherited from pretraining, the task information provided by demonstrations, and the corrections learned through autonomous interaction.

\vspace{-0.1cm}
\section{Preliminaries}

As defined in \cref{sec:problem_setting}, the MDA regime seeks to learn a policy through autonomous interaction with a target environment $\Mtarget$. The objective of this autonomous learning is to solve the target task consistently, and in as few environment interactions as possible. We formalize this objective in the standard reinforcement learning (RL) setting.
Concretely, the goal is to learn a policy $\pi: \mathcal{S} \to \Delta(\mathcal{A})$ that maximizes the discounted return $J(\pi;\mathcal{M}) = \mathbb{E}^{\mathcal{M},\pi}[\sum_{t=0}^{\infty} \gamma^t r(s_t,a_t)]$, where $s_0 \sim \rho$, $a_t \sim \pi(\cdot \mid s_t)$, and $s_{t+1} \sim P(\cdot \mid s_t,a_t)$. We focus on sparse binary rewards, where $r_t=0$ if the task is completed within the time limit and $-1$ otherwise. Because the policy incurs a penalty of $-1$ at every step until the task is solved, maximizing this cumulative return implicitly optimizes for solving the task consistently and in the fewest environment interactions, thereby recovering our original formulation of autonomous learning. A central object utilized by our approach is the action-value function $Q^\pi(\bs,\ba)$, which estimates the expected discounted return from taking action $\ba$ in state $\bs$ and following $\pi$ thereafter. In our robotic setting, states include high-dimensional visual and proprioceptive observations, and actions are chunks~\citep{zhao2023learningfinegrainedbimanualmanipulation}: $\ba_t$ denotes a sequence of low-level controls executed open-loop. Details are in \cref{app:action-chunking}.

\textbf{Value-Based RL.} Value-based methods estimate $Q^\pi$ from experience: they maintain a replay buffer $\mathcal{B}$ of previously collected transitions and train a critic using temporal-difference (TD) learning. Given a transition  $(\bs_t,\ba_t,r_t,\bs_{t+1}) \sim \mathcal{B}$, the TD target and loss are
\begin{align}
    \iftoggle{arxiv}{}{\textstyle} y_t
    &=
    r_t
    +
    \gamma Q_{\bar{\psi}}(\bs_{t+1},\ba'),
    \notag\\
    \mathcal{L}_{\mathrm{TD}}(\psi)
    &=
    \mathbb{E}_{\mathcal{B}}
    \left[
        \left(
            Q_\psi(\bs_t,\ba_t) - y_t
        \right)^2
    \right],
\label{eq:td_loss}
\end{align}
where $Q_{\bar{\psi}}$ is a target critic and $\ba'$ is the next action. The policy $\pi_\theta$ is then updated toward actions that receive high value under the critic \cite{fujimoto2018addressingfunctionapproximationerror,haarnoja2018softactorcriticoffpolicymaximum}.

\textbf{Residual RL.}
Residual RL adapts a frozen base policy by learning a lightweight module that modifies the base policy's proposed action~\citep{dong2026expostablereinforcementlearning, xu2026rltokenbootstrappingonline, ankile2024imitationrefinementresidual, ankile2025residualoffpolicyrlfinetuning}.
Given a policy $\pi_{\theta}$, the residual policy $\pi_{\theta}^{\mathrm{res}}$ conditions on both the state and an action proposed by the base policy, and outputs the action executed in the environment, i.e. $\ba_{\mathrm{exec}} \sim \pi_{\theta}^{\mathrm{res}}(\cdot \mid \bs, \ba_{\mathrm{base}})$ with $\ba_{\mathrm{base}} \sim \pi_{\theta}(\cdot \mid \bs)$.
This residual update can be parameterized in several ways, such as predicting an additive offset in action space or directly predicting the modified action.
In this work, we adopt the latter parameterization as discussed in \cref{sec:midas}.

\textbf{Policy-Agnostic RL.}
Policy-agnostic RL (PA-RL)~\citep{mark2024policyagnosticrloffline} is an online improvement method that applies across policy classes.
It first improves sampled actions against the learned Q-function, then distills the optimized actions back into the parametric policy using the appropriate supervised objective, such as flow matching for flow policies or likelihood for discrete-action policies.
We use PA-RL for this policy-agnostic structure, extending it to action-chunked residual policies (details in \cref{app:parl_details}).
For each replay state $\bs_t$, we sample candidate action chunks from the residual policy and refine them by taking gradient steps against $Q_\psi$,
\begin{align}
    \iftoggle{arxiv}{}{\textstyle} \ba_t
    &\sim
    \pi_{\theta}^{\mathrm{res}}(\cdot \mid \bs_t),
    \notag\\
    \ba_t^\star
    &\leftarrow
    \ba_t + \eta \, \nabla_{\ba} Q_\psi(\bs_t, \ba_t),
    \label{eq:parl_refine}
\end{align}
where $\eta$ is the step size, and then distill the optimized chunk $\ba_t^\star$ back into the residual policy.

\textbf{Generalist robot policy.}
We take the base policy $\pibase$ to be a generalist robot policy, such as a pretrained vision-language-action model like $\pifive$~\citep{physicalintelligence2025pi05}. Concretely, $\pibase$ consists of a pretrained vision-language-model (VLM) backbone and a flow head that parameterizes a flow policy conditioned on the output of the VLM backbone.

\section{\midas: A Simple Recipe for Minimal Data Adaptation}
\label{sec:midas}


In this section, we propose \textbf{Mi}nimal-\textbf{D}ata \textbf{A}daptation \textbf{S}trategy (\midas), a simple representative recipe for adapting a pretrained policy from minimal demonstrations and autonomous interaction.
 We use \midas to  provide a common experimental framework for examining the roles of demonstrations, pretrained representations, and online corrective learning through controlled comparisons and targeted ablations. \midas builds prior work on residual and offline-to-online RL methods~\citep{ zhou2025efficientonlinereinforcementlearning, ball2023efficientonlinereinforcementlearning}, but introduces subtle-yet-important modifications to how past methods utilize interaction data so as to account for \emph{minimal} pretrained success. These serve to expose the differences between MDA and the traditional, large-BC-success offline-to-online regime.  We note that concurrent work \citep{sun2026priorproefficientskill} introduces alternative recipes that are similar in spirit and motivation.

The main motivation for \midas is that fine-tuning on a small number of $K$ demonstrations seeds coarse task-following behavior but does not ensure reliable completion (shown in \Cref{sec:what_enables_midas}).
This ``coarse'' behavior following provides a stronger prior than using pretraining solely for representation learning \citep{shah2021rrlresnetrepresentationreinforcement, parisi2022unsurprisingeffectivenesspretrainedvision, kim2024openvlaopensourcevisionlanguageactionmodel,zhou2025dinowmworldmodelspretrained}, but is a weaker prior in that pretraining does not attain non-trivially large task completion success, as is the common regime in robot learning and language modeling \citep{hu2024imitationbootstrappedreinforcementlearning, Guo_2025}.

In our intermediate regime,  we find that learning a lightweight residual policy through value-based RL often suffices to close the remaining gap.
These findings directly lead to \midas: a simple two-stage recipe combining few-demo behavior cloning with online residual RL. We denote the resulting adapted policy by $\pi_{\mathrm{RL}}^K$ and provide the full algorithm in \cref{app:algorithm}.
We note that the two-stage structure follows established approaches to initializing policies from offline data and improving them through online interaction
\citep{zhou2025efficientonlinereinforcementlearning,ball2023efficientonlinereinforcementlearning}. Our focus here is on how to use \emph{minimal offline} data to complete a given task.

\textbf{Stage I: Behavior cloning on minimal demonstrations.} Stage I adapts the pretrained policy $\pibase$ to the downstream task using the $K$ demonstrations in $\Ddemo^K$.  We first fine-tune the policy on the available demonstrations to obtain a task-directed initialization for subsequent online RL.
Implementation details are deferred to \cref{app:bc_details}. As described in \Cref{sec:what_enables_midas}, the resulting  $\pibase^K$  coarsely attempts the task, but does not complete it reliably.

\textbf{Stage II: Online improvement via value-based residual RL.} Stage II improves $\pibase^K$ through autonomous interaction using value-based RL.
We instantiate this adaptation module as a residual policy conditioned on the action proposed by $\pibase^K$, following prior work ~\citep{dong2026expostablereinforcementlearning, xu2026rltokenbootstrappingonline, ankile2024imitationrefinementresidual, ankile2025residualoffpolicyrlfinetuning} with $\pibase^K$ kept frozen. At each state $\bs_t$, the base policy proposes an action chunk $\ba_t^{\mathrm{base}}$, and the residual policy $\pi_\theta^{\mathrm{res}}$ proposes an edit conditioned on the state and the chunk itself. We parameterize the residual policy as follows \footnote{Unlike typical residual formulations, \eqref{eq:residual} directly predicts the edited action $\ba_t^{\mathrm{exec}} \sim \pi_\theta^{\mathrm{res}}(\cdot | \bs_t,\ba_t^{\mathrm{base}})$. } (further details on design choices in \cref{app:residual_parameterization})
\begin{equation}
  \label{eq:residual}
      \pires(\cdot \mid \bs_t, \abase)
      =
      \tanh\!\left(
          \mathcal{N}\!\left(
              \mu_\theta(\bs_t,\abase),\;
              \sigma_\theta^2(\bs_t,\abase)
          \right)
      \right).  
       \end{equation}

\subsection{Overcoming The Challenges Introduced By Minimal Demonstrations}
The two-stage structure alone does not ensure reliable learning in MDA. With only a few demonstrations, the residual actor and critic have little data for initialization, and autonomous rollouts from $\pibase^K$ may succeed only 2--5\% of the time. Moreover, the initial policy may not reliably produce the corrective actions needed for task completion (\Cref{sec:what_enables_midas}). Our residual PA-RL implementation addresses these challenges through the initialization, replay sampling, and policy-update choices described below.

\textbf{1) Offline warmup:  residual actor initialization and critic calibration from few demonstrations.} A randomly initialized residual actor would produce actions unrelated to the base policy, making early exploration unstable.
We therefore construct a ``warmup'' buffer $\mathcal{B}_{\mathrm{warm}}$ from the $K$ demonstrations in $\Ddemo^K$ together with a small number of autonomous rollouts from $\pibase^K$ that may or may not be successful. For each state in this buffer, we query the frozen base policy to obtain $\ba^{\mathrm{base}}\sim\pibase^K(\cdot\mid\bs)$, and train the residual actor to reproduce this base proposal:
{\small
  \vspace{-0.5em}
  \begin{align}
      \mathcal{L}_{\mathrm{warm}}(\theta)
      =
      \mathbb{E}_{(\bs,\ba^{\mathrm{base}})\sim\mathcal{B}_{\mathrm{warm}},
      \;\ba\sim\pi_\theta^{\mathrm{res}}(\cdot\mid\bs,\ba^{\mathrm{base}})}
      \left[
          \left\|
          \ba-\ba^{\mathrm{base}}
          \right\|_2^2
      \right].
      \label{eq:residual_bc_warmup}
  \end{align}
  \vspace{-2em}
  }

This initialization makes the residual actor approximately preserve the behavior of $\pibase^K$:
Online RL also requires a critic that can guide improvement from the start. While prior work can pretrain such critics from offline data~\citep{kumar2022ptr,nakamoto2024calqlcalibratedofflinerl, zhou2025efficientonlinereinforcementlearning, kostrikov2021offlinereinforcementlearningimplicit}, our setting has no additional offline data beyond the few demonstrations and autonomous rollouts from $\pibase^K$. We therefore draw inspiration from \citet{zhou2025efficientonlinereinforcementlearning} and train the critic on the warmup buffer $\mathcal{B}_{\mathrm{warm}}$ using TD learning (Eq.~\eqref{eq:td_loss}). Importantly, this critic is trained on the state-action distribution induced by $\pibase^K$, making it calibrated to the regions where online improvement begins and limiting forgetting.

\textbf{2) Success balancing: overcoming limited initial policy performance.} Since the base policy may only succeed 2-5\% in some cases after Stage I,
we employ \emph{success balancing} to amplify the sparse reward signal during both the warmup and online improvement phases. To do so, we maintain a replay buffer $\mathcal{B}$ containing all collected trajectories and a success buffer $\mathcal{B}_{\mathrm{succ}}$ containing only successful trajectories, both initialized with the given $K$ demonstrations.
Critic minibatches are sampled from a mixture of these buffers, oversampling $\mathcal{B}_{\mathrm{succ}}$ to prevent the value function from being dominated by failed rollouts and to guide policy improvement toward reward-bearing regions. Details are in \cref{app:success_buffer}.

\textbf{3) Residual policy training with PA-RL~\citep{mark2024policyagnosticrloffline}.}
Lastly, \Cref{sec:what_enables_midas} reveals the need for actions that go beyond the support of the base policy. While many value-based RL algorithms can provide this, we opt for ``policy-agnostic'' PA-RL approach,  which we apply  to the residual actor. Having a policy-agnostic parameterization allows us to easily extend our method to more sophisticated residual policies like a minimal-iterative-policy \citep{pan2026adonoisingdispellingmyths}, while also being competitive with other common policy extraction approaches \cite{haarnoja2018softactorcriticoffpolicymaximum, fujimoto2018addressingfunctionapproximationerror}.
We implement this by sampling \emph{a single} proposal action $\abase_t$ from the base policy, and then applying  PA-RL style Best-of-N distillation objective with an action gradient to a batch of  $N$ actions $\ba^{(i)}_t \sim \pires(\cdot \mid \bs_t, \abase_t)$ sampled from the residual (See \cref{app:parl_details} for more details). Importantly, sampling requires  \emph{only} a single expensive forward pass through $\pibase^K$ while sampling the remaining actions from the lightweight residual policy for diversity.

\enlargethispage{2\baselineskip}
\begin{AIbox}{\midas{}: Key design decisions}
\begin{itemize}[leftmargin=6mm,itemsep=2pt,topsep=4pt]
    \item \textbf{Two-stage recipe: LoRA BC, then residual RL.} Stage~I performs LoRA behavior cloning on the $K$ demonstrations
  $\Ddemo^K$ to obtain a coarse prior $\pibase^K$; Stage~II freezes $\pibase^K$ and runs sparse-reward online RL on a
lightweight residual actor-critic on top of it.
    \item \textbf{Three ingredients make Stage~II work.} \textbf{(1)} an \emph{offline warmup} for calibrating the residual actor and critic; \textbf{(2)} \emph{success balancing} the batch for critic updates; and \textbf{(3)} \emph{PA-RL residual updates} that push beyond the
base policy's support.
\end{itemize}
\end{AIbox}

\section{MDA is Possible From As Little As One Demonstration}
\label{sec:performance}

\begin{table}[t]
\centering
\scriptsize
\renewcommand{\arraystretch}{0.92}
\setlength{\tabcolsep}{2pt}

\resizebox{0.9\columnwidth}{!}{%
\begin{tabular}{l c ccccc}
\multicolumn{7}{c}{\textbf{LIBERO-Long}} \\[2pt]
\toprule
& & \multicolumn{5}{c}{\textbf{1 demonstration}} \\
\cmidrule(lr){3-7}
Task & Zero-Shot & BC & DSRL & Filt.\ BC & DiceRL & \textbf{\midas} \\
\midrule
\quad Alph.\ Soup + Cr.\ Cheese $\to$ Basket
& 0.0 & 12.7\pms{1.2} & 12.0\pms{2.0} & 14.0\pms{2.0} & 98.0\pms{2.0} & \textbf{99.3\pms{1.2}} \\
\quad Alph.\ Soup + Tom.\ Sauce $\to$ Basket
& 0.0 & 4.0\pms{3.5} & 3.3\pms{2.3} & 2.0\pms{0.0} & \textbf{92.0\pms{4.0}} & 82.7\pms{13.3} \\
\quad Black Bowl $\to$ Bottom Drawer
& 0.0 & 22.7\pms{6.4} & 28.7\pms{4.2} & 28.7\pms{10.3} & 90.7\pms{3.1} & \textbf{98.7\pms{1.2}} \\
\quad Book $\to$ Caddy
& 0.0 & 37.3\pms{7.0} & 42.7\pms{6.1} & 28.0\pms{2.0} & 98.0\pms{3.5} & \textbf{98.7\pms{2.3}} \\
\quad Both Moka Pots $\to$ Stove
& 0.0 & 13.3\pms{3.1} & 14.7\pms{4.6} & 19.3\pms{7.0} & 74.7\pms{9.5} & \textbf{76.0\pms{14.0}} \\
\quad Cr.\ Cheese + Butter $\to$ Basket
& 0.0 & 34.0\pms{9.2} & 28.0\pms{2.0} & 68.0\pms{2.0} & 88.7\pms{2.3} & \textbf{96.0\pms{0.0}} \\
\quad Moka Pot $\to$ Stove
& 0.0 & 84.0\pms{0.0} & 92.0\pms{0.0} & 85.3\pms{8.3} & \textbf{98.7\pms{1.2}} & 95.3\pms{3.1} \\
\quad White Mug + Choc.\ Pudding
& 0.0 & 33.3\pms{11.7} & 45.3\pms{6.4} & 42.0\pms{4.0} & \textbf{97.3\pms{4.6}} & 96.7\pms{5.8} \\
\quad White Mug + Plates
& 0.0 & 30.0\pms{2.0} & 17.3\pms{1.2} & 34.7\pms{6.4} & 61.3\pms{11.5} & \textbf{79.3\pms{11.0}} \\
\quad Yellow-White Mug $\to$ Microwave
& 0.0 & 64.0\pms{2.0} & 50.7\pms{4.6} & 68.0\pms{7.2} & \textbf{93.3\pms{3.1}} & 89.3\pms{8.1} \\
\midrule
\textbf{Average}
& 0.0 & 33.5\pms{1.8} & 33.5\pms{0.8} & 39.0\pms{2.4} & 89.3\pms{1.8} & \textbf{91.2\pms{4.2}} \\
\bottomrule

\end{tabular}%
}

\vspace{6pt}

\resizebox{0.9\columnwidth}{!}{%
\begin{tabular}{l c ccccc}
\multicolumn{7}{c}{\textbf{RoboCasa}} \\[2pt]
\toprule
& & \multicolumn{5}{c}{\textbf{1 demonstration}} \\
\cmidrule(lr){3-7}
Task & Zero-Shot & BC & DSRL & Filt.\ BC & DiceRL & \textbf{\midas} \\
\midrule
\quad Banana: Fridge Drawer $\to$ Shelf
& 0.0 & 13.3\pms{3.1} & 12.7\pms{2.3} & 19.3\pms{3.1} & \textbf{95.3\pms{1.2}} & 93.3\pms{7.0} \\
\quad Hot Dog: Counter $\to$ Cabinet
& 0.0 & 29.3\pms{1.2} & 20.0\pms{10.4} & 36.0\pms{29.5} & 93.3\pms{6.1} & \textbf{100.0\pms{0.0}} \\
\quad Mug $\to$ Coffee Machine + Start
& 0.0 & 24.0\pms{8.7} & 24.7\pms{4.2} & 46.7\pms{7.6} & 19.3\pms{33.5} & \textbf{74.7\pms{20.0}} \\
\quad Cup + Bowl $\to$ Dishwasher + Close$^{\dagger}$
& 0.0 & 0.0\pms{0.0} & 0.0\pms{0.0} & 0.0\pms{0.0} & 0.0\pms{0.0} & 0.0\pms{0.0} \\
\midrule
\textbf{Average}
& 0.0 & 22.2\pms{3.4} & 19.1\pms{3.2} & 34.0\pms{10.6} & 69.3\pms{13.3} & \textbf{89.3\pms{8.7}} \\
\bottomrule
\end{tabular}%
}
\caption{\textbf{One demonstration adaptation results measured over 3 seeds.} Success rates (\%) are mean $\pm$ std. $^{\dagger}$Excluded from the RoboCasa average.}
\label{tab:one_demo_results}
\vspace{-1em}
\end{table}

In this section, we evaluate representative adaptation strategies in the MDA regime and show that successful adaptation is possible from as little as one demonstration.

\textbf{Experimental setup.}
We use $\pifive$~\citep{physicalintelligence2025pi05} as the pretrained base policy $\pibase$ and run \midas with $K{=}1$ demonstration only on two simulation benchmarks: \textbf{a)} LIBERO-Long \citep{liu2023liberobenchmarkingknowledgetransfer}, which studies long-horizon language-conditioned manipulation, and \textbf{b)} RoboCasa-365 \citep{nasiriany2026robocasa365largescalesimulationframework}, which studies long-horizon household manipulation in diverse kitchen scenes (Details in Appendix \ref{app:sim_details}).

\textbf{Experimental Results.} \cref{tab:one_demo_results} compares representative strategies corresponding to the three ingredients of adaptation in MDA: pretraining alone (\emph{zero-shot}), adaptation with a single demonstration (\emph{BC}), and subsequent online improvement. For online improvement, we evaluate learning from successful rollouts (\emph{Filtered BC}), value-guided noise-space steering (\emph{DSRL}~\citep{wagenmaker2025steeringdiffusionpolicylatent}), and value-guided action-space adaptation with \midas.
We also compare with \emph{DICE-RL}~\citep{sun2026priorproefficientskill}, which is a concurrent method that adapts frozen generative policies with value-based residual corrections and selective behavior regularization. DICE-RL shares key design principles underlying \midas (see \cref{app:dicerl_shared_principles}). This allows us to test whether successful adaptation extends to alternative implementations of our shared principles. We see that \midas and DICE-RL recover strong performance from one-demonstration initializations on LIBERO, however on the longer-horizon (Mug $\to$ Coffee Machine + Start) task on Robocasa, DICE-RL substantially underperforms \midas. We hypothesize that this difference arises from DICE-RL not backing up online successes separately in the success buffer, which becomes crucial on longer horizon tasks to amplify the sparse reward signal. Filtered BC and DSRL provide smaller and less consistent improvements, particularly when initial success is below $20\%$. 
Implementation details in \cref{app:sim_details}.

\textbf{Design choices affect online improvement.} We examine the effects of replay sampling and continued behavior-cloning regularization in \Cref{app:ablations}. We find that removing success balancing lowers performance. Early in training, most autonomous rollouts fail, so uniform replay sampling provides relatively few examples of successful execution. Success balancing keeps these rare successful transitions present in actor and critic updates. However, explicitly encouraging the actor to imitate successful trajectories throughout online learning also reduces performance, suggesting that continued imitation can restrict useful departures from previously observed actions.
These ablations highlight the importance of different design choices in successful adaptation with \midas.

\textbf{Stage I is necessary for bootstrapping sparse-reward online learning.}
We ablate Stage I by running the full \midas online learning procedure directly from the zero-shot base policy. This variant fails to learn, achieving $0\%$ success throughout training. Stage I is therefore critical for moving the policy into a task-relevant region of behavior from which sparse-reward online learning can bootstrap.

\section{What Enables Minimal-Data Adaptation?}

\label{sec:what_enables_midas}
Previously, we motivated minimal-data adaptation with the premise that a few demonstrations can recover enough task-directed behavior to make online improvement tractable. In this section, we present detailed empirical evidence examining how pretraining, demonstrations, and online interaction enable successful adaptation in this regime.
First, the few-demonstration BC finetuning of the pretrained VLA provides actions which follow task instructions coarsely, reducing the burden of exploration.
Second, the pretrained VLA provides visual representations, enabling efficient learning without direct state access.  Finally, online reinforcement fills in the gaps in precise-control, which necessarily requires value-guided online updates to finetuning actions beyond  what $\pibase^K$ can reliably produce on its own. Additional ablation experiments for \midas are deferred to Appendix~\ref{app:ablations} and experiments to Appendix~\ref{app:baselines}.

\begin{figure}[t]
    \centering
    \includegraphics[width=\columnwidth]{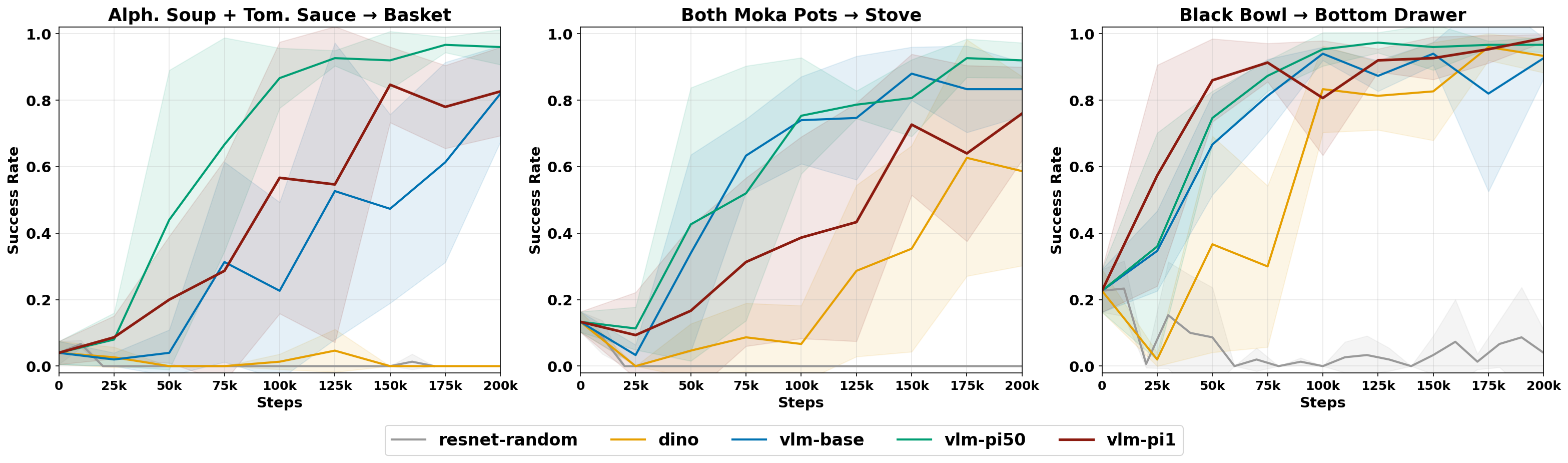}
    \captionsetup{font=footnotesize}
    \caption{\textbf{Performance of different representations in online adaptation.} We see that (a) No pretraining, in the form of a Resnet backbone fine-tuned from scratch, collapses; (b) Generic visual pretraining, in the form of frozen DINO features helps, but takes longer to converge; (c) Pretrained VLA representations of $\pi_\text{0.5}$ are highly sample efficient, consistent with the hypothesis that action-aligned pretraining can substantially improve the sample efficiency of downstream online adaptation.}
    \label{fig:libero_reps}
    \vspace{-1em}
\end{figure}
\begin{wrapfigure}{h}{0.34\columnwidth}
    \centering
    \includegraphics[width=\linewidth]{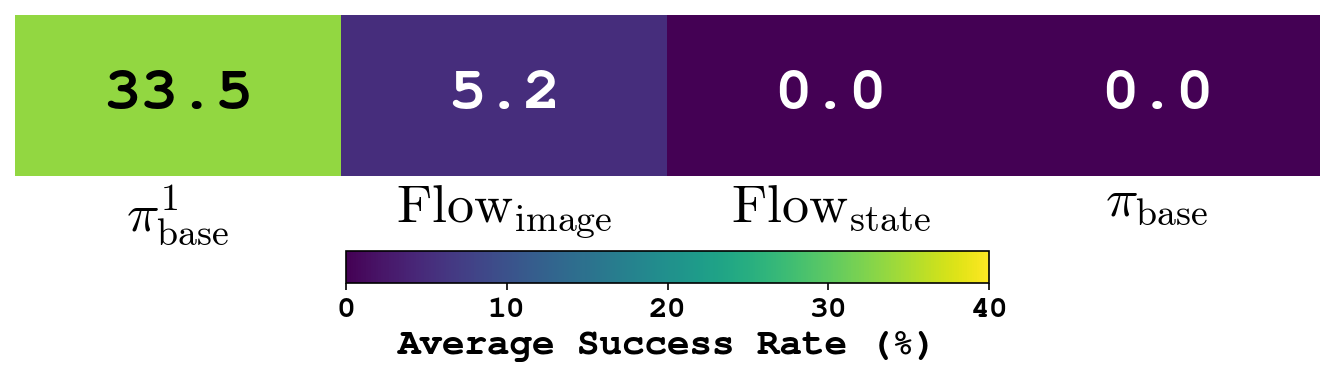}
    \vspace{-0.3cm}
    \caption{\footnotesize{
    \textbf{Performance of different state representations for behavior cloning}
    }}
    \label{fig:dp_bc_cmp}
    \vspace{-0.5cm}
\end{wrapfigure}

\textbf{1) Pretraining and demonstrations enable learning of coarse task-specific behaviors:}
Before online adaptation, we ask whether $K{=}1$ can recover coarse task behavior.
Figure~\ref{fig:dp_bc_cmp} compares $\pibase^1$ to three approaches: \textbf{a)} a flow-matching policy trained on pretrained ResNet features, which tests whether generic visual pretraining is sufficient; \textbf{b)} a flow-matching policy trained on privileged state inputs, which tests whether near-optimal state information alone enables recovery of task relevant behavior; and \textbf{c)} the zero-shot pretrained policy $\pibase$, which tests whether pretraining alone identifies the task. All three perform poorly: the ResNet flow-policy achieves non-zero success on only 3/10 LIBERO-Long tasks, while the privileged-state policy and $\pibase$ obtain 0\% success. Together, these failures suggest that MDA requires both task grounding from the demonstration and behavior-aware representations from pretraining; privileged state inputs alone do not recover this behavior, suggesting that the benefit of pretraining goes beyond improved state representation.

\begin{figure}[t]
    \centering
    \includegraphics[width=\columnwidth]{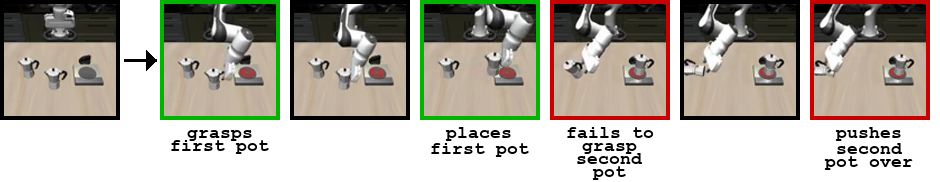}
    \captionsetup{font=footnotesize}
    \caption{\textbf{Few-demo BC recovers task-level behavior, but not fine-grained control.}
    Rollouts of $\pibase^K$ at $K{=}1$ on \emph{Both Moka Pots $\to$ Stove}. The policy reaches the right objects but fails to grasp the second pot under orientation changes of the environment's reset distribution.}
    \label{fig:bc_rollouts}
    \vspace{-1em}
\end{figure}

Qualitative rollouts in Fig.~\ref{fig:bc_rollouts} show what $\pibase^1$ recovers despite its low success. The policy often approaches the correct object, attempts the correct subtask sequence, and follows a trajectory resembling the demonstration. Its failures are primarily in reliable execution: contact, alignment, grasping, and recovery remain brittle. Thus, pretraining and one demonstration learn coarse task behavior, but online adaptation is needed for reliable control.

\textbf{2) Pretrained representations simplify online adaptation and substantially improve sample efficiency:}
The preceding results show that large-scale pretraining, together with minimal demonstrations, anchors the search for task-relevant behavior. We now find that the representations provided to the residual policy also determine how efficiently online RL can exploit this initialization.
Figure~\ref{fig:libero_reps} isolates this effect by varying the representation used by the residual learner.
We compare \textbf{(i)} a \textbf{ResNet} trained from scratch during online RL, representing no visual pretraining; \textbf{(ii)} frozen \textbf{DINO} features \citep{caron2021emergingpropertiesselfsupervisedvision}, representing generic large-scale self-supervised visual pretraining; and \textbf{(iii)} representations from the pretrained VLA backbone. Within the VLA representations, we further compare $\pibase$, $\pibase^1$, and $\pibase^{50}$ to assess the effect of increasing amounts of demo data.

\begin{wrapfigure}{r}{0.5\columnwidth}
    \centering
    \begin{subcaptionbox}{Trajectory critical state\label{fig:umap_critical_state_a}}[0.22\columnwidth]
        {\includegraphics[width=0.22\columnwidth]{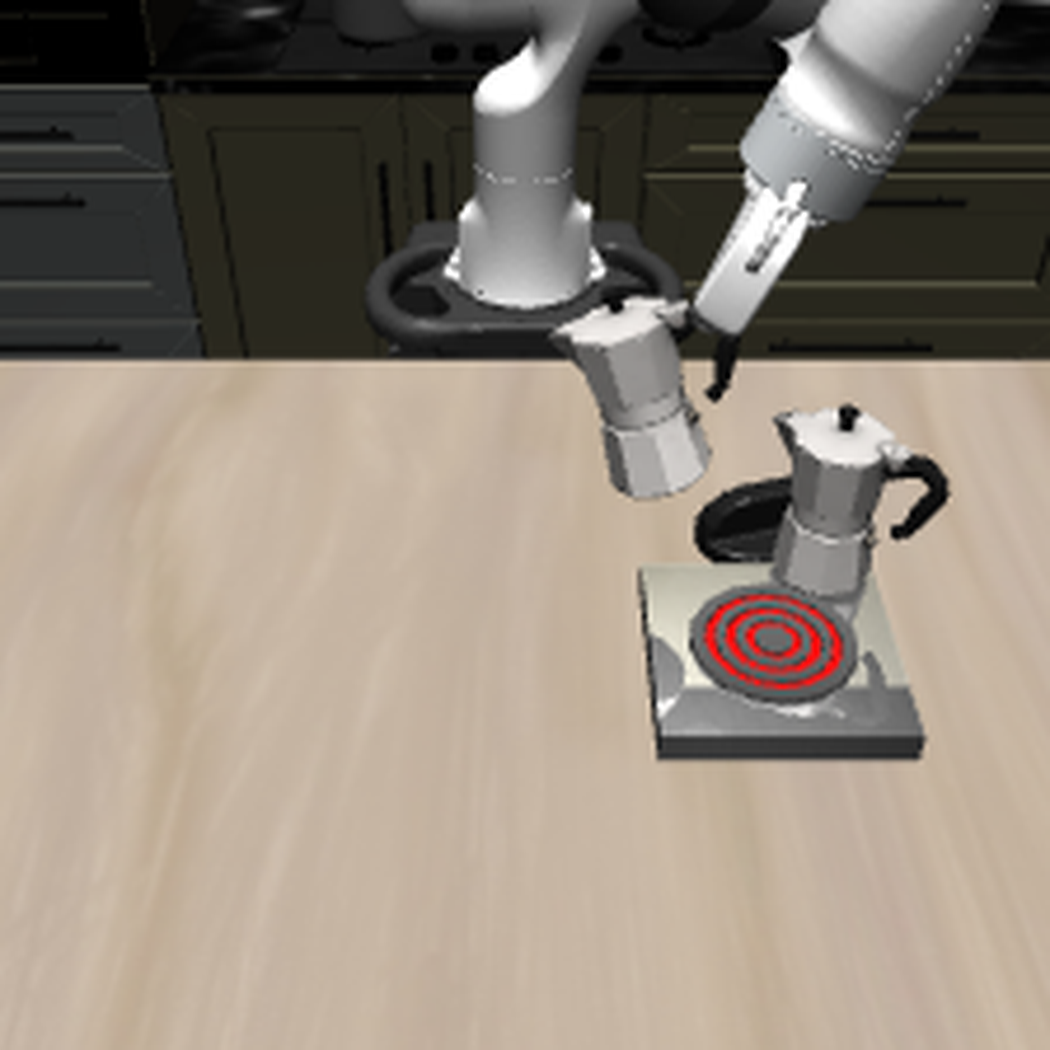}}
    \end{subcaptionbox}
    \hspace{4pt}
    \begin{subcaptionbox}{UMAP of actions\label{fig:umap_critical_state_b}}[0.22\columnwidth]
        {\includegraphics[width=0.22\columnwidth]{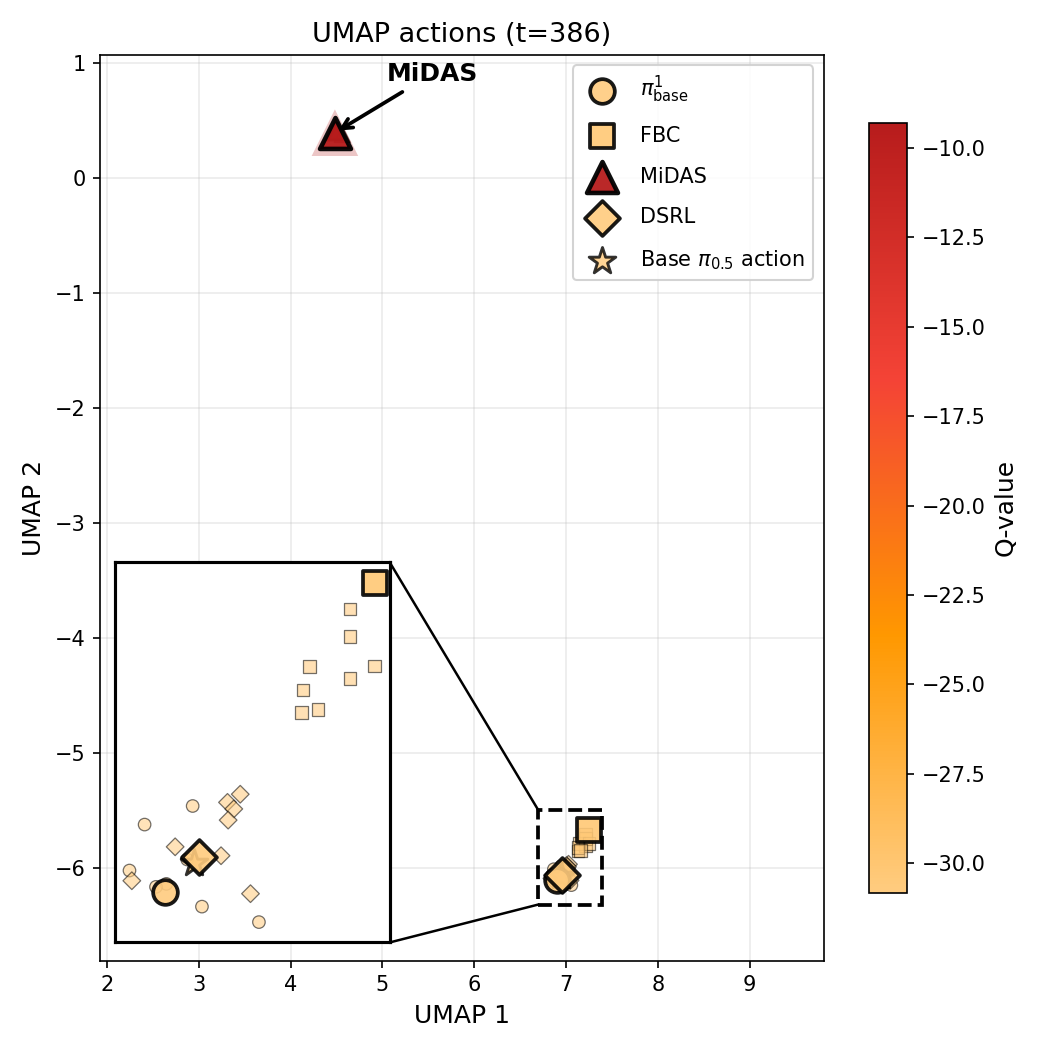}}
    \end{subcaptionbox}
    \captionsetup{font=footnotesize}
    \caption{\textbf{$\midas$ reaches actions outside the effective support of $\pibase^K$.}
    Filtered BC and DSRL remain near samples from $\pibase^K$ and fail; $\midas$ selects an action in a disjoint region with higher $Q^\ast$ and succeeds.}
\label{fig:umap_critical_state}
\end{wrapfigure}

Without visual pretraining, \midas fails to converge at $K=1$, indicating that sparse interaction is insufficient to simultaneously learn task-relevant visual features and residual action corrections. Generic visual pretraining with frozen DINO substantially improves learning, but still requires considerably more interaction than VLA features, suggesting that generic visual invariances alone are insufficient for sample-efficient adaptation. In contrast, representations from the pretrained $\pi_{0.5}$ backbone enable rapid improvement throughout training. The similar performance of $\pibase, \pibase^1$ and $\pibase^{50}$ on two of the three tasks indicates that task-specific fine-tuning provides little additional improvement to the representations used for online adaptation.
Although we study a single VLA, these results suggest that action-aligned representations from  pretraining can substantially improve the sample efficiency of online adaptation.

\textbf{3) Value-based RL learns fine-grained control  actions beyond the base policy's  prior:} 
The previous analysis showed that $\pibase^K$ often produces task-relevant behavior, but solves the task inconsistently. 
\begin{wraptable}[9]{l}{0.45\columnwidth}
    \centering
    \resizebox{\linewidth}{!}{%
    \begin{tabular}{lcc}
        \toprule
        \textbf{Method} & \textbf{\#Samples} & \textbf{Min. Dist. to Base Cloud} \\
        \midrule
        Base  & 33 & 0.0142 (within cloud) \\
        FBC   & 33 & 0.1122 \\
        DSRL  & 33 & 0.0244 \\
        MiDAS & 33 & 4.5371 \\
        \bottomrule
    \end{tabular}%
    }
    \caption{\footnotesize{
    \textbf{Minimum Distance of action clouds of various methods from the base action cloud.}
    }}
    \label{tab:base_cloud_distance}
\end{wraptable}
We ask: do the right actions already lie within its support but are not reliably selected, or do they fall outside its effective support? We test this by comparing $\midas$ to the baselines introduced in \cref{sec:performance}.
\renewcommand{\pms}[1]{{\tiny$\pm$#1}}
We examine a ``critical'' state in a LIBERO-Long rollout in Fig.~\ref{fig:umap_critical_state} to make the distinction between \midas and the baselines. At this critical state, actions from Filtered BC and DSRL remain concentrated near samples from $\pibase^1$ and fail, while $\midas$ selects an action in a disjoint region of the action space and succeeds.


To check that the UMAP visualization reflects separation in the robot's original chunked action space, we also compute distances between the action samples. In \cref{tab:base_cloud_distance}, we see that the $L_{2}$ distance of actions from $\pibase^{1}$ of \midas is substantially larger compared to that of DSRL and Filtered-BC, validating the UMAP results \cref{fig:umap_critical_state}. These observations corroborate the hypothesis that the successful action does not lie within the base policy's effective support;
sharpening or steering within this support cannot reach it. Breaking out of this support, therefore, requires online adaptation that can expand the action distribution beyond the effective support of $\pibase^1$ and \midas is able to do so by learning value-guided corrections.

\begin{figure}[!t]
    \centering
     \includegraphics[width=0.85\columnwidth]{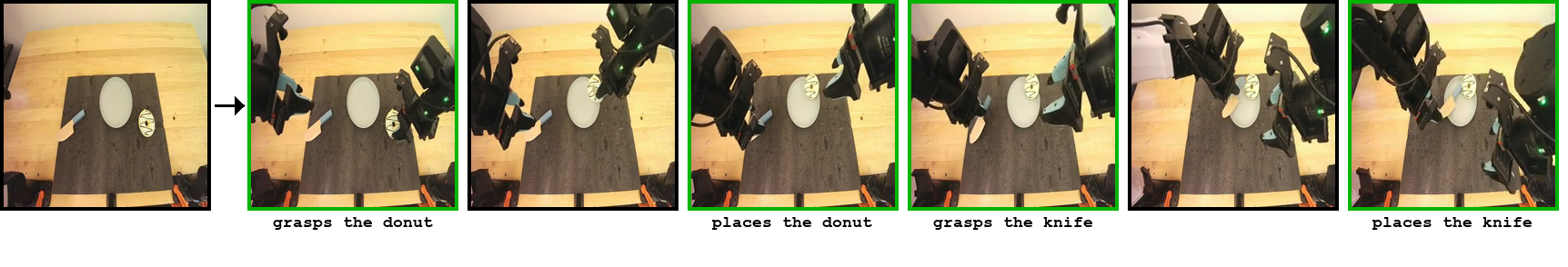}    \includegraphics[width=0.85\columnwidth]{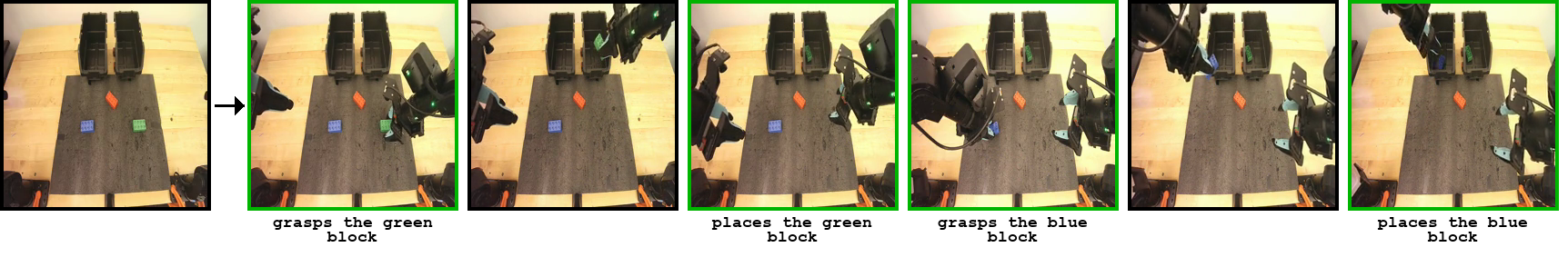}
    \captionsetup{font=footnotesize}
    \vspace{-0.5cm}
    \caption{\textbf{Representative real-world rollouts.} Rollouts illustrating task progress on the knife-and-donut task (top) and the block placement task (bottom).}
    \label{fig:qualitative_filmstrips}
\end{figure}

\label{sec:real_world}
\begin{wraptable}{r}{0.28\columnwidth}
    \centering
    \vspace{-1em}
    \caption{\textbf{One demonstration real-world adaptation results.}
    Success rates (\%) for BC and \midas{} over 15 rollouts.}
    \label{tab:real_world_results}
    \vspace{-0.5em}
    \scriptsize
    \setlength{\tabcolsep}{2pt}
    \renewcommand{\arraystretch}{0.9}
    \begin{tabular}{@{}lcc@{}}
        \toprule
        & \multicolumn{2}{c}{\textbf{1 Demo}} \\
        \cmidrule(lr){2-3}
        Task & BC & \textbf{\midas} \\
        \midrule
        \textit{Real World} & & \\
        \quad Block Placement
            & 40.0 & \textbf{67.0} \\
        \quad Knife + Donut $\to$ Plate
            & 27.0 & \textbf{80.0} \\
        \midrule
        \textbf{Average} & 33.5 & \textbf{73.5} \\
        \bottomrule
    \end{tabular}
    \vspace{-1em}
\end{wraptable}
\textbf{4) Minimal-data adaptation is feasible on real hardware.}
We finally evaluate whether minimal-data adaptation is tractable in the real world using a bimanual YAM platform. We consider two pick-and-place tasks: \textbf{(T1)} placing a green block in the right container and a blue block in the left container, and \textbf{(T2)} placing a knife and a donut on a plate (Figure \ref{fig:qualitative_filmstrips}). In both tasks, we use $\pifive$~\citep{physicalintelligence2025pi05} as $\pibase$ and apply the same \midas protocol with a single demonstration ($K{=}1$); implementation details are deferred to Appendix \ref{app:real_world_exps}. On \textbf{T1}, the one demonstration policy $\pibase^1$ already reaches $40\%$ success, which we hypothesize reflects overlap with the pretraining distribution of $\pifive$. Online RL further improves success to $67\%$ and produces qualitatively new corrective behaviors not observed in base-policy rollouts (Appendix \ref{app:real_world_exps}). On the harder \textbf{T2} task, where the warm-start policy is substantially weaker, online adaptation improves success from $27\%$ to $80\%$ after $5$-$6$ hours of autonomous interaction. 



\begin{AIbox}{Key findings}
\begin{itemize}[leftmargin=4mm,itemsep=2pt]
    \item \textbf{Expert demonstrations induce coarse task-solving behavior.} Behavior cloning on minimal demonstrations instills rough task solving  behavior into the pretrained policy.
    \item \textbf{Pretrained representations give strong priors for adaptation.} The VLM-head's features, used as the state abstraction for control and value learning, enable rapid online adaptation.
    \item \textbf{Value-based RL expands the base policy's action support.} PA-RL extraction from an online-learned value function adds fine-grained corrections beyond the base policy's action support.
\end{itemize}
\end{AIbox}

\section{Generalization and Robustness in Minimal Data Adaptation}
\label{sec:generalization}

The preceding analysis examined how pretraining, demonstrations, and online interaction enable minimal-data adaptation. We now study how these capabilities extend beyond the conditions encountered during adaptation: what robustness is inherited from pretraining, what generalization online learning enables, and which shifts remain difficult for the adapted policy.

We consider two categories of shifts: \textbf{(a) observation shifts}, which alter object appearance (color, texture) or language phrasing while preserving the task and its physical requirements, and \textbf{(b) state shifts}, which change dynamically relevant quantities such as starting positions, object shapes, or object instances. We ask: \textbf{(1)} does adaptation preserve robustness to visual and language changes? \textbf{(2)} Under which state shifts does online improvement transfer? and \textbf{(3)} can broader online experience improve robustness to larger state shifts without additional demonstrations?


As in Section~\ref{sec:what_enables_midas}, we evaluate on the 10 long-horizon tasks in LIBERO-Long. For visual, shape, and object-instance shifts, we use the perturbation definitions from LIBERO-PRO~\citep{zhou2026liberoprorobustfairevaluation}. For position shifts, we perturb target objects at three levels: low (5\,cm / 0.1\,rad), medium (10\,cm / 0.15\,rad), and high (20\,cm / 0.25\,rad). (Further details in \cref{app:sim_perturb_details})

\begin{figure}[!t]
    \centering
    \includegraphics[width=\columnwidth]{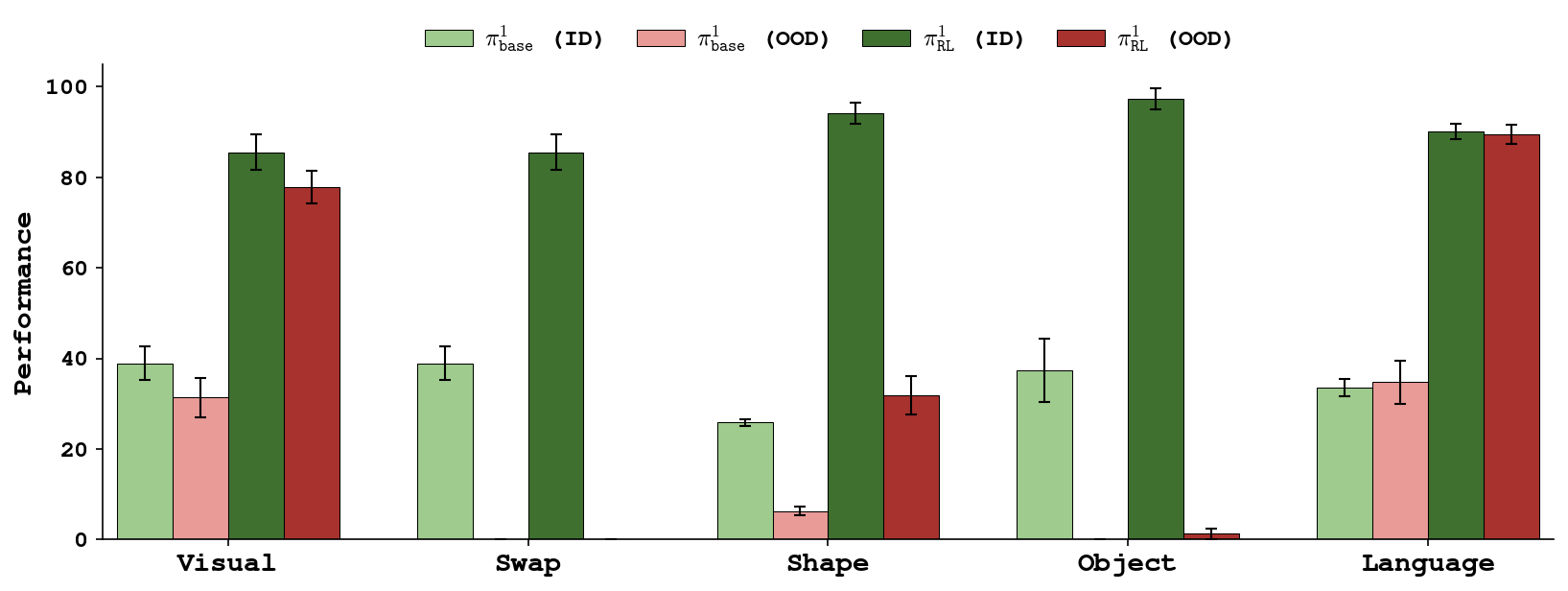}
    \captionsetup{font=footnotesize}
    \caption{\footnotesize{Performance of BC and online RL under observation and state shifts. \textbf{Observation Shifts (Visual/Language)}: Pretrained representations preserve performance under visual and language changes. \textbf{State Shifts (Swap/Shape/Object)}: Online RL improves robustness to moderate shifts but performance degrades under larger shifts.}}
    \label{fig:v2_perturbation_comparison}
\end{figure}

\textbf{(Q1) Observation-level robustness is inherited from the frozen vision-language-model (VLM) backbone.}
We first test whether minimal-data adaptation preserves robustness to semantic input changes, rather than merely fitting the visual and language conditions seen in the demonstrations. Under visual shifts such as color and texture changes (Fig.~\ref{fig:v2_perturbation_comparison}), $\pibase^K$ ($K{=}1$) retains most of its in-distribution performance, and the adapted policy $\pi_{\mathrm{RL}}^1$ shows a similar retention pattern despite using only a lightweight residual actor. The same holds under language paraphrases (Fig.~\ref{fig:v2_perturbation_comparison}). Since both policies share the same frozen VLM backbone, and the RL module adds little perceptual or linguistic capacity, this robustness is best explained by invariances already present in the pretrained representation. Thus, online RL improves control while retaining observation-level robustness.

\textbf{(Q2) Generalization to state shifts depends on whether new behavior is required.} We next evaluate shifts that require the policy to change its actions.
Figure~\ref{fig:v2_perturbation_comparison}, shows that $\pi_{\mathrm{RL}}^1$ improves over $\pibase^1$ under some state shifts, especially shape changes where the required grasp is similar. This result can simply be explained by the fact that
the policy just needs to execute the same behavior in a different part of the state space. However, performance under both drops sharply under object swaps and object changes: changes in object affordance or placement can require distinct manipulation strategies, which are difficult to infer from one demonstration and online interaction around it.
\cref{tab:state_shift_perturb} summarizes these results. (Further details in \cref{app:sim_perturb_exps}).

\begin{wrapfigure}{r}{0.42\columnwidth}
    \centering
    \vspace{-3em}
    \captionsetup{font=footnotesize}
    \includegraphics[width=0.4\columnwidth]{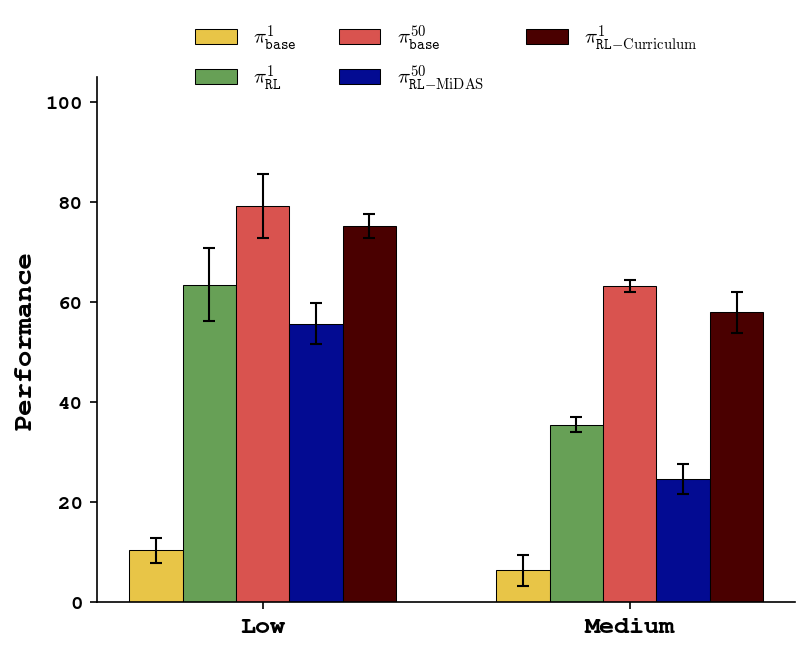}
    \caption{\textbf{Curriculum closes the gap.} Progressively widening the reset distribution during online RL recovers most of the position-generalization gap with $\pibase^{50}$ without additional demos.}
    \label{fig:v2_curriculum_result}
    \vspace{-2em}
\end{wrapfigure}

\textbf{(Q3) Broader coverage, either from demonstrations or curriculum, expands the generalization boundary.}  Lastly, we ask: can autonomous adaptation ensure robustness to large perturbation shifts that are outside the support of the reset distribution, provided that they do not require fundamentally new behavior (in view of Q2)? 

In \Cref{app:position_generalization_analysis}, we show that naively increasing the breadth of demonstration can improve robustness. In \Cref{app:curriculum_gen}, we describe a curriculum that gradually increases the width of the distribution, and achieves similar robustness to a policy trained on $50$ demonstrations. \cref{fig:v2_curriculum_result} summarizes our results and shows that by gradually widening the reset distribution with $\pi^1_{\text{RL-Curriculum}}$, we are able to recover most of the performance of $\pi^{50}_{\text{base}}$ under large perturbation shifts.

\begin{AIbox}{Generalization of policies learned in MDA}
\begin{itemize}[leftmargin=4mm,topsep=2pt,itemsep=2pt]
  \item \midas generalizes to \textbf{observation shifts} via the frozen VLM backbone and to \textbf{state shifts requiring no new
behavior} via online RL.
  \item Shifts demanding new grasps or strategies remain out of reach, but a \textbf{reset-distribution
curriculum} extends coverage to larger position and orientation shifts.
\end{itemize}
\end{AIbox}

\section{Related Work}
\label{app:related_work}

\begin{table}[!t]
      \centering
      \begin{subtable}[t]{0.60\textwidth}
      \centering
      \resizebox{\linewidth}{!}{%
      \begin{tabular}[t]{l cc cc}
      \toprule
       & \multicolumn{2}{c}{$\pibase^1$} & \multicolumn{2}{c}{$\pi_{\mathrm{RL}}^1$} \\
      \cmidrule(lr){2-3} \cmidrule(lr){4-5}
      Task & ID & OOD & ID & OOD \\
      \midrule
      Both Moka Pots $\to$ Stove & 13.3\pms{3.1} & 0.0\pms{0.0} & 76.0\pms{14.0} & 0.0\pms{0.0} \\
      Cr.\ Cheese + Butter $\to$ Basket & 34.0\pms{9.2} & 0.0\pms{0.0} & 96.0\pms{0.0} & 0.0\pms{0.0} \\
      Moka Pot $\to$ Stove & 84.0\pms{0.0} & 0.0\pms{0.0} & 95.3\pms{3.1} & 0.0\pms{0.0} \\
      White Mug + Choc.\ Pudding & 33.3\pms{11.7} & 0.0\pms{0.0} & 96.7\pms{5.8} & 0.0\pms{0.0} \\
      White Mug + Plates & 30.0\pms{2.0} & 0.0\pms{0.0} & 79.3\pms{11.0} & 0.0\pms{0.0} \\
      \midrule
      \textbf{Average} & 38.9\pms{3.6} & 0.0\pms{0.0} & 88.7\pms{5.9} & 0.0\pms{0.0} \\
      \bottomrule
      \end{tabular}}
      \subcaption{Object swap.}
      \label{tab:swap_perturb}
      \end{subtable}
      \hfill
      \begin{minipage}[t]{0.375\textwidth}
      \begin{subtable}[t]{\linewidth}
      \centering
      \resizebox{\linewidth}{!}{%
      \begin{tabular}[t]{l cc cc}
      \toprule
       & \multicolumn{2}{c}{$\pibase^1$} & \multicolumn{2}{c}{$\pi_{\mathrm{RL}}^1$} \\
      \cmidrule(lr){2-3} \cmidrule(lr){4-5}
      Task & ID & OOD & ID & OOD \\
      \midrule
      Black Bowl $\to$ Bottom Drawer & 22.7\pms{6.4} & 20.7\pms{6.4} & 98.7\pms{1.2} & 90.7\pms{8.1} \\
      Yellow-White Mug $\to$ Microwave & 64.0\pms{2.0} & 4.7\pms{3.1} & 89.3\pms{8.1} & 20.7\pms{13.3} \\
      Alph.\ Soup + Cr.\ Cheese $\to$ Basket & 12.7\pms{1.2} & 0.0\pms{0.0} & 99.3\pms{1.2} & 2.7\pms{4.6} \\
      Alph.\ Soup + Tom.\ Sauce $\to$ Basket & 4.0\pms{3.5} & 0.0\pms{0.0} & 82.7\pms{13.3} & 2.0\pms{3.5} \\
      \midrule
      \textbf{Average} & 25.8\pms{0.8} & 6.3\pms{1.0} & 92.5\pms{5.4} & 29.0\pms{5.8} \\
      \bottomrule
      \end{tabular}}
      \subcaption{Shape change.}
      \label{tab:shape_perturb}
      \end{subtable}


      \begin{subtable}{\linewidth}
      \centering
      \resizebox{\linewidth}{!}{%
      \begin{tabular}{l cc cc}
      \toprule
       & \multicolumn{2}{c}{$\pibase^1$} & \multicolumn{2}{c}{$\pi_{\mathrm{RL}}^1$} \\
      \cmidrule(lr){2-3} \cmidrule(lr){4-5}
      Task & ID & OOD & ID & OOD \\
      \midrule
      \makebox[0pt][l]{Book $\to$ Caddy}\hphantom{Alph.\ Soup + Cr.\ Cheese $\to$ Basket} & 37.3\pms{7.0} & 0.0\pms{0.0} &
  98.7\pms{2.3} &
       2.0\pms{2.0} \\
      \bottomrule
      \end{tabular}}
      \subcaption{Object change.}
      \label{tab:obj_change_perturb}
      \end{subtable}
      \end{minipage}
      \caption{\textbf{Per-task success rate (\%) under state-shift perturbations.}
      \textbf{(a)} Object swap: exchanging the initial positions of two scene objects drops both $\pibase^1$ and $\pi_{\mathrm{RL}}^1$
   to
      0\% despite strong ID performance.
      \textbf{(b)} Shape change: replacing the target with a geometrically different variant of the same category preserves
  performance
      when grasp affordances transfer, but degrades sharply when the required contact geometry changes.
      \textbf{(c)} Object change: replacing the target with a categorically different object defeats both policies, as the new
  affordances
       require manipulation strategies outside the adapted policy's behavioral support.}
      \label{tab:state_shift_perturb}
\end{table}

\textbf{Generalist robot policies and VLAs.}
Large-scale robot pretraining has enabled generalist policies that can condition on language and visual observations to perform diverse manipulation tasks. Prior work has introduced transformer-based robot policies and vision-language-action models trained on broad robot and web-scale data, including RT-1~\citep{brohan2023rt1roboticstransformerrealworld}, RT-2~\citep{brohan2023rt2visionlanguageactionmodelstransfer}, Octo~\citep{octomodelteam2024octoopensourcegeneralistrobot}, $\pi_0$~\citep{black2024pi0}, $\pifive$~\citep{physicalintelligence2025pi05}, world-action models~\citep{ye2026worldactionmodelszeroshot}, and efficient VLA fine-tuning methods~\citep{kim2025openvlaoft}. These models provide reusable perceptual representations and broad action priors, making them natural candidates for fast downstream adaptation. However, despite this progress, current generalist policies are still not reliable zero-shot learners in new environments: they may fail to reach sparse rewards or visit too narrow a set of task-relevant states for autonomous learning to bootstrap. In contrast, we study whether such pretrained policies can be made adaptable from minimal task-specific supervision.

\textbf{Few-shot robot imitation and adaptation.}
A complementary line of work studies learning from very small numbers of demonstrations. One-shot and few-shot imitation learning methods~\citep{duan2017oneshotimitationlearning, finn2017oneshotvisualimitationlearning} show that robots can infer new tasks from one or a few examples, often through meta-learning or in-context task conditioning~\citep{dipalo2024keypointactiontokensenable}. Other approaches study rapid adaptation from short demonstrations~\citep{haldar2023teachrobotfishversatile} or in-context imitation policies~\citep{vosylius2025instantpolicyincontextimitation}. While appealing in theory, few shot and in-context imitations suffer from two structural limitations. First, context-conditioned policies are known to suffer form causal confusion or are prone to shortcut features \citep{de2019causal}, and evidence from large-language-model literature shows that task-specific fine-tuning often outperforms in-context learning \citep{liu2022fewshotparameterefficientfinetuningbetter}. Second, and more fundamentally, these methods do not have any mechanism for autonomously improving from their own experience. Motivated by these two observations, we study the MDA regime. With very few demonstrations for specifying the task, the MDA regime uses them as a scaffold to make autonomous interaction tractable and analyzes the components enabling this tractability.

\textbf{Online and residual RL for robot policy adaptation.}
Online reinforcement learning and offline-to-online methods provide tools for improving robot policies after deployment. Prior work has studied offline RL pretraining~\citep{kumar2022ptr}, calibrated offline-to-online fine-tuning~\citep{nakamoto2024calqlcalibratedofflinerl}, efficient online RL with offline data~\citep{ball2023efficientonlinereinforcementlearning}, imitation-bootstrapped RL~\citep{hu2024imitationbootstrappedreinforcementlearning}, and policy-agnostic RL for fine-tuning diverse policy classes~\citep{mark2024policyagnosticrloffline}. More closely related, recent methods use RL to improve expressive policies or pretrained robot policies, including diffusion-policy steering~\citep{wagenmaker2025steeringdiffusionpolicylatent}, diffusion policy optimization~\citep{ren2024diffusionpolicypolicyoptimization}, RL-token adaptation~\citep{xu2026rltokenbootstrappingonline}, distribution-contractive fine-tuning~\citep{sun2026priorproefficientskill}, and residual RL for refining imitation policies~\citep{dong2026expostablereinforcementlearning, ankile2024imitationrefinementresidual, ankile2025residualoffpolicyrlfinetuning}. 
These methods provide foundations for improving pretrained robot policies through online interaction. 

Building on this body of existing work, we study \emph{minimal-data adaptation}: whether one or a few demonstrations can make autonomous improvement tractable when the pretrained policy is initially unreliable. 
Our emphasis is on isolating how pretraining, demonstrations, and online experience jointly enable adaptation, which algorithmic choices make learning effective in this regime, and what forms of generalization emerge or remain limited. From this study, we propose \midas,   a representative practical recipe built on prior offline-to-online and residual RL approaches. \textbf{\textit{Concurrent work,}} DICE-RL~\citep{sun2026priorproefficientskill}, improves frozen robot policies by learning value-based residual corrections, and combining demonstration data with online experience. 
DICE-RL shares key design principles underlying \midas and its success on several MDA tasks (see \cref{sec:performance}) provides further evidence for the shared principles underlying MiDAS in enabling successful adaptation. Our comparisons also highlight the importance of specific algorithmic choices within this framework, with MiDAS achieving stronger performance on a longer-horizon RoboCasa task. Our contribution lies in establishing and analyzing the MDA regime, understanding the principles that enable adaptation, and identifying design choices that make learning effective under minimal task-specific supervision.


\section{Discussion and Future Work}
\label{app:future_work}

Future work should study how to make exploration more effective. Our results show that a single demonstration can anchor exploration around a narrow tube of task-relevant behavior, within which autonomous RL can improve reliably. However, learning qualitatively new behaviors or reaching distinct behavioral modes remains substantially harder. Expanding the reset distribution provides one mechanism for broadening exploration, but our curriculum results also show that performance degrades as this distribution widens, and that reset expansion cannot recover behaviors that lie outside the support of the initial policy. This suggests that autonomous improvement will require multiple forms of supervision to work together, including demonstrations, reset curricula, language-based steering, and reward signals. Future work can characterize how these sources can be combined to discover new behavior while balancing human effort, computational cost, interaction efficiency, and final-policy robustness.

Second, MDA provides a setting for understanding what demonstrations are actually needed for. Given one demonstrated behavior, online RL can robustify and extend that behavior over a surrounding region of the task distribution, but there is an upper bound to what can be recovered from this initialization alone. This raises the question of when additional demonstrations become necessary: whether one demonstration is required for each distinct behavioral mode, what constitutes a mode in robot control, and which variations RL can generalize across without further task-specific supervision. The answer may also depend on how the demonstration is incorporated. We use behavior cloning followed by a memory-efficient residual policy, which may not extract all of the behavior that one trajectory makes available. More expressive adaptation, including larger task-specific modules or full-policy fine-tuning, may make better use of the same demonstration, but introduces different tradeoffs in compute, memory, stability, and sample efficiency.

Third, it remains important to determine how broadly the MDA regime extends beyond the behaviors studied here. Our experiments primarily consider pick-and-place tasks that are plausibly well represented in the pretraining distribution of $\pi_{0.5}$ (and other generalist policies of today). It is therefore unclear whether the same interaction between pretraining, minimal demonstrations, and autonomous RL will hold when adaptation requires greater policy capacity or more substantial behavioral change. Future work should test this regime across new embodiments and grippers, dexterous and precision-critical manipulation, assembly, tool use, and longer-horizon tasks. These settings can reveal whether the present findings reflect a general property of pretrained robot policies or a regime that is currently limited to behaviors already well supported by pretraining.

Finally, MDA exposes a distinction between memorizing task-specific behavior and generalizing across deployment variation. Our results suggest that different components of the pipeline support different forms of robustness: pretrained visual representations provide invariance to some observation-level changes, while demonstrations and online interaction determine which task-relevant states and behaviors the policy can reliably handle. Autonomous interaction therefore does not imply broader generalization by itself; the resulting robustness depends on the distribution of states encountered and the behaviors already available to the policy. Future work should characterize which forms of generalization arise from pretraining, which must be induced through task-specific demonstrations or online interaction, and how these components should be balanced to obtain policies that both adapt efficiently and generalize beyond the conditions encountered during adaptation.

\vspace{-0.2cm}
\section*{Acknowledgements}
We would like to thank Shashwat Saxena, 
Zheyuan Hu, and Sarvesh Patil for informative discussions and Naveen Enock for assistance with the hardware setup. SV thanks Soumojit Bhattacharya and Max Sobol Mark for writing feedback and research discussions. The authors thank the members of the CMU Maxlab and CMU AIRe lab for feedback and support. AK acknowledges funding from Office of Naval Research under grant number N00014-24-1-2206 and support from Toyota Research Institute U3.0 project. MS acknowledges support from Toyota Research Institute U3.0. We thank the Babel cluster at CMU and TPU resources from Google Cloud for computing support.

\bibliographystyle{plainnat}
\bibliography{refs}
\newpage

\appendix
\crefalias{section}{appendix}
\crefalias{subsection}{appendix}
\crefname{appendix}{Appendix}{Appendices}
\Crefname{appendix}{Appendix}{Appendices}

\part*{Appendices}


Additional videos of rollouts and training are available here: \href{https://minimal-data-adaptation.github.io/}{minimal-data-adaptation.github.io}.

\section{Action chunking}
\label{app:action-chunking}
We apply action chunking \citep{zhao2023learningfinegrainedbimanualmanipulation}, where
sequences of actions $u_{t:t+C-1}$ are predicted and executed in open-loop.
Let $u_t \in \R^d$ denote a primitive robot action and let $C$ denote the chunk
horizon. We define the action space used by the policy as
$\mathcal{A} \subseteq \R^{C \times d}$, so that each action
$a_t \in \mathcal{A}$ corresponds to an entire action chunk,
$a_t = (u_t,\ldots,u_{t+C-1})$. After sampling
$a_t \sim \pi(\cdot \mid s_t)$, the robot executes the chunk open-loop for $C$
environment steps before replanning from the resulting state $s_{t+C}$. For notational
simplicity, we treat each chunk as a single action in the MDP, preserving the
standard MDP notation above.

When using chunked actions, the reward associated with a transition is the
discounted reward accumulated over the executed chunk,
and bootstrapping occurs from $s_{t+C}$ with discount $\gamma^C$. Thus, unless stated otherwise, $a_t$ denotes an action chunk and $r_t$ denotes the reward accumulated over the executed chunk, with bootstrapping discount $\gamma^C$. For our experiments, we use a reward of -1 for a chunk with effective discount of $\gamma^C$.

\section{Residual Policy Parameterization}
\label{app:residual_parameterization}

We parameterize the residual actor by conditioning it on $\ba_t^{\mathrm{base}}$. This exposes it directly to the action mode selected by $\pibase^K$, so online RL refines a task-relevant VLA proposal rather than learning to represent a complex, multimodal action distribution from scratch. At the same time, predicting $\ba_t^{\mathrm{exec}}$ directly allows the policy to shift this proposal toward higher-value regions instead of being restricted to small fixed residual corrections around $\ba_t^{\mathrm{base}}$ that further limit the policy and are more common in the literature\citep{dong2026expostablereinforcementlearning}.
Concretely, an additive parameterization would predict a bounded, additive correction $\ba_t^{\mathrm{exec}} = \ba_t^{\mathrm{base}} + \Delta_\theta(\bs_t,\ba_t^{\mathrm{base}})$, constraining the executed action to a fixed neighborhood of the base proposal. Because minimal demonstrations do not guarantee that near-optimal actions lie close to $\ba_t^{\mathrm{base}}$ (\Cref{sec:what_enables_midas}), such a bounded residual can prevent the policy from reaching higher-value actions. Predicting $\ba_t^{\mathrm{exec}}$ directly instead lets the base proposal anchor exploration while leaving the policy free to move beyond its immediate neighborhood. Crucially, the amount of movement around the proposed action can now be heterogeneous across states, whereas the additive approach would bound the movement by a fixed constant at all states.
We then evaluate this action by a critic ensemble $\{Q_{\psi_j}(\bs_t,\ba_t^{\mathrm{exec}})\}_{j=1}^{N}$ trained with temporal difference learning in Eq.~\eqref{eq:td_loss}. We discuss implementation details for this residual parameterization in \cref{app:parl_details}.

For the state representation $s_t$, we use the output of the VLM head by taking a mean across all token positions. The residual actor is parameterized as a 3-layer MLP with a hidden dimension of 512 and ReLU activation.  

\typeout{DEBUG: residual_parameterization.tex was loaded}

\section{Algorithm}
\label{app:algorithm}

\newcounter{algbox}
\crefname{algbox}{Algorithm}{Algorithms}
\Crefname{algbox}{Algorithm}{Algorithms}

Algorithm~\ref{alg:midas} summarizes \midas using the residual policy in
\cref{eq:residual} and action chunks of length $H=C$ as in
\cref{app:action-chunking}. Hyperparameters used below are listed at the top
of the algorithm; additional implementation settings are in
\cref{tab:parl_hyperparams}.

\phantomsection\label{app:algo_replay}
\textbf{Replay and sampling.}
Store $(\bs_t,\ba_t^{\mathrm{base}},\ba_t^{\mathrm{exec}},r_t,m_t,\bs_{t+H})$,
where $\bs_t=(\phi(o_t),q_t)$ combines mean-pooled VLM prefix features
$\phi(o_t)$ of observation $o_t$ and proprioception $q_t$.
Set $(r_t,m_t)=(0,0)$ for successful terminal chunks and
$(-1,1)$ otherwise; demonstrations use their recorded actions.
The operator $\operatorname{clip}$ clips each action to the action bounds.

\phantomsection\label{app:algo_updates}
\textbf{Update conventions.}
TD updates use \cref{eq:td_loss} with ensemble-mean target values,
next actions sampled from $\pires(\cdot\mid\bs',\ba'^{\mathrm{base}})$,
where $\ba'^{\mathrm{base}}$ is the frozen base policy's proposal at the next state.
Use discount $\gamma^H m_t$, clip the sampled next action, and update targets
after every critic update via $\bar\psi\gets\tau\psi+(1-\tau)\bar\psi$.
All TD and BC minibatches have size $B$; optimizer settings are in
\cref{app:parl_details}.

\newcommand{\algln}[2]{%
  \hangindent=2.6em
  \makebox[1.9em][r]{\footnotesize#1}\hspace{0.7em}#2\par
}

\begin{figure}[tp]
\centering
\refstepcounter{algbox}\label{alg:midas}%
\begin{tcolorbox}[enhanced, arc=1.5pt,
  colback=white, colframe=black, boxrule=0.7pt,
  left=6pt, right=6pt, top=4pt, bottom=5pt,
  fonttitle=\bfseries, coltitle=white, colbacktitle=black,
  title={Algorithm \thealgbox\quad\midas: Minimal-Data Adaptation Strategy}]
\footnotesize
\setlength{\parindent}{0pt}
\setlength{\parskip}{1.6pt}

\textbf{Require:}~Pretrained policy $\pibase$, $K$ successful demonstrations
$\Ddemo^K$, and target environment (\cref{sec:problem_setting})\par
\textbf{Ensure:}~Adapted policy $\pi_{\mathrm{RL}}^K$\par

\textbf{Hyperparameters:}\par
\emph{Replay and critics:} chunk length $H$, batch size $B$,
number of critics $J$, success fraction $\rho_{\mathrm{succ}}$,
discount $\gamma$, Polyak rate $\tau$.\par
\emph{Schedule:} demo warmup rounds per transition $U_{\mathrm{demo}}$,
warmup threshold $W$ rounds, minimum replay size $n_{\min}$,
update budget $T_{\max}$ rounds;
critic/actor updates per round:
warmup $(U_C^{\mathrm{w}},U_A^{\mathrm{w}})$,
RL $(U_C,U_A)$.\par
\emph{PA-RL:} candidates $N$, elites $M$, ascent steps $G$,
ascent step size $\eta$, actor learning rate $\alpha_\pi$.\par

\vspace{2pt}
\textit{Stage I: Behavior cloning on $K$ demonstrations}\par
\algln{1}{Fine-tune $\pibase$ on $\Ddemo^K$ with flow matching
  $\rightarrow\pibase^K$ (\cref{app:bc_details})}

\vspace{1pt}
\textit{Stage II: Residual reinforcement learning}\par
\algln{2}{Freeze $\pibase^K$; initialize actor $\pires$ (\cref{eq:residual}),
  critics $\{Q_{\psi_j}\}_{j=1}^{J}$ (\cref{app:parl_details}),
  and targets $\bar\psi\gets\psi$}
\algln{3}{Define the mean critic
  $\bar Q(\bs,\ba)\coloneqq J^{-1}\sum_{j=1}^{J}Q_{\psi_j}(\bs,\ba)$}
\algln{4}{Use the state, transition, and clipping definitions on
  p.~\pageref{app:algo_replay} and the TD/BC rules on p.~\pageref{app:algo_updates}}
\algln{5}{Initialize replay $\mathcal{B}$ and success replay $\mathcal{B}_{\mathrm{succ}}$
  with all stride-one demo transitions and base proposals sampled from $\pibase^K$}
\algln{6}{Run $U_{\mathrm{demo}}|\mathcal{B}|$ warmup rounds:
  $U_C^{\mathrm{w}}$ TD (\cref{eq:td_loss}) and
  $U_A^{\mathrm{w}}$ BC (\cref{eq:residual_bc_warmup}) updates,
  sampling uniformly from $\mathcal{B}$}
\algln{7}{Set $i\gets U_{\mathrm{demo}}|\mathcal{B}|$, the number of completed update rounds}
\algln{8}{\textbf{while} $i\le T_{\max}$ \textbf{do}}
\algln{9}{\hspace{1.2em}Collect an episode in $H$-step chunks (\cref{app:action-chunking}):
  at each chunk, sample
  $\ba_t^{\mathrm{base}}\sim\pibase^K(\cdot\mid o_t)$ with fresh flow-matching noise}
\algln{10}{\hspace{2.4em}Execute the clipped base proposal if $i<W$;
  otherwise the clipped deterministic action $\tanh\!\left(\mu_\theta(\bs_t,\ba_t^{\mathrm{base}})\right)$}
\algln{11}{\hspace{1.2em}Add transitions to $\mathcal{B}$;
  if the episode succeeded, also add them to $\mathcal{B}_{\mathrm{succ}}$}
\algln{12}{\hspace{1.2em}\textbf{if} $|\mathcal{B}|\le n_{\min}$:
  \textbf{continue}}
\algln{13}{\hspace{1.2em}Define $\mathcal{B}_{\mathrm{mix}}$ as the mixture of the current
  $\mathcal{B}$ and $\mathcal{B}_{\mathrm{succ}}$ in \cref{eqn:success_buffer}}
\algln{14}{\hspace{1.2em}\textbf{for} each chunk collected in the episode \textbf{do}}
\algln{15}{\hspace{2.4em}$(u_C,u_A)\gets(U_C^{\mathrm{w}},U_A^{\mathrm{w}})$
  if $i<W$, else $(U_C,U_A)$}
\algln{16}{\hspace{2.4em}Perform $u_C$ TD updates (\cref{eq:td_loss})
  with minibatches from $\mathcal{B}_{\mathrm{mix}}$}
\algln{17}{\hspace{2.4em}\textbf{if} $i<W$: perform $u_A$ BC updates
  (\cref{eq:residual_bc_warmup}) on uniform minibatches from $\mathcal{B}$}
\algln{18}{\hspace{2.4em}\textbf{else}: perform $u_A$ PA-RL updates as follows}
\algln{19}{\hspace{3.6em}Sample minibatch $\mathcal{D}\sim\mathcal{B}_{\mathrm{mix}}$,
  $|\mathcal{D}|=B$; for each $(\bs,\ba^{\mathrm{base}})\in\mathcal{D}$:}
\algln{20}{\hspace{4.8em}Sample $\{\ba^{(n)}\}_{n=1}^{N}$ using
  \cref{eq:residual}, conditioned on $(\bs,\ba^{\mathrm{base}})$}
\algln{21}{\hspace{4.8em}$\mathcal{C}\gets
  \{\operatorname{clip}(\ba^{\mathrm{base}})\}\cup
  \{\operatorname{clip}(\ba^{(n)})\}_{n=1}^{N}$}
\algln{22}{\hspace{4.8em}$\mathcal{E}\gets$ the $M$ elements of $\mathcal{C}$
  with highest $\bar Q(\bs,\ba)$}
\algln{23}{\hspace{4.8em}Refine each $\ba\in\mathcal{E}$ for $G$ steps using
  \cref{eq:parl_refine} with $\bar Q$; clip after each step}
\algln{24}{\hspace{4.8em}$\ba^\star(\bs,\ba^{\mathrm{base}})\gets
  \argmax_{\ba\in\mathcal{E}}\bar Q(\bs,\ba)$;
  stop gradients through this target}
\algln{25}{\hspace{3.6em}$\mathcal{L}_{\mathrm{PA}}(\theta)\gets
  \mathbb{E}_{\substack{(\bs,\ba^{\mathrm{base}})\sim\mathcal{D}\\
    \hat\ba\sim\pires(\cdot\mid\bs,\ba^{\mathrm{base}})}}
  \big[\|\hat\ba-\ba^\star(\bs,\ba^{\mathrm{base}})\|_2^2\big]$}
\algln{26}{\hspace{3.6em}Update $\theta$ with reparameterized
  $\nabla_\theta\mathcal{L}_{\mathrm{PA}}$: Adam at $\alpha_\pi$
  (\cref{app:parl_details})}
\algln{27}{\hspace{2.4em}\textbf{end if}}
\algln{28}{\hspace{2.4em}$i\gets i+1$}
\algln{29}{\hspace{1.2em}\textbf{end for}}
\algln{30}{\textbf{end while}}
\algln{31}{\textbf{return} $\pi_{\mathrm{RL}}^K$:
  sample $\ba^{\mathrm{base}}\sim\pibase^K$ and execute $\tanh\!\left(\mu_\theta(\bs_t,\ba_t^{\mathrm{base}})\right)$}
\end{tcolorbox}
\end{figure}

\section{Simulation Experiment Details}
\label{app:sim_details}
We use the Libero and RoboCasa Benchmarks for our simulation experiments

\subsection{Libero and RoboCasa Task Details}

We evaluate on two simulation benchmarks: LIBERO-Long~\citep{liu2023liberobenchmarkingknowledgetransfer} and RoboCasa-365~\citep{nasiriany2026robocasa365largescalesimulationframework}. Below we describe each benchmark and the specific tasks used.

\begin{figure}[t]
    \centering
    \includegraphics[width=\linewidth]{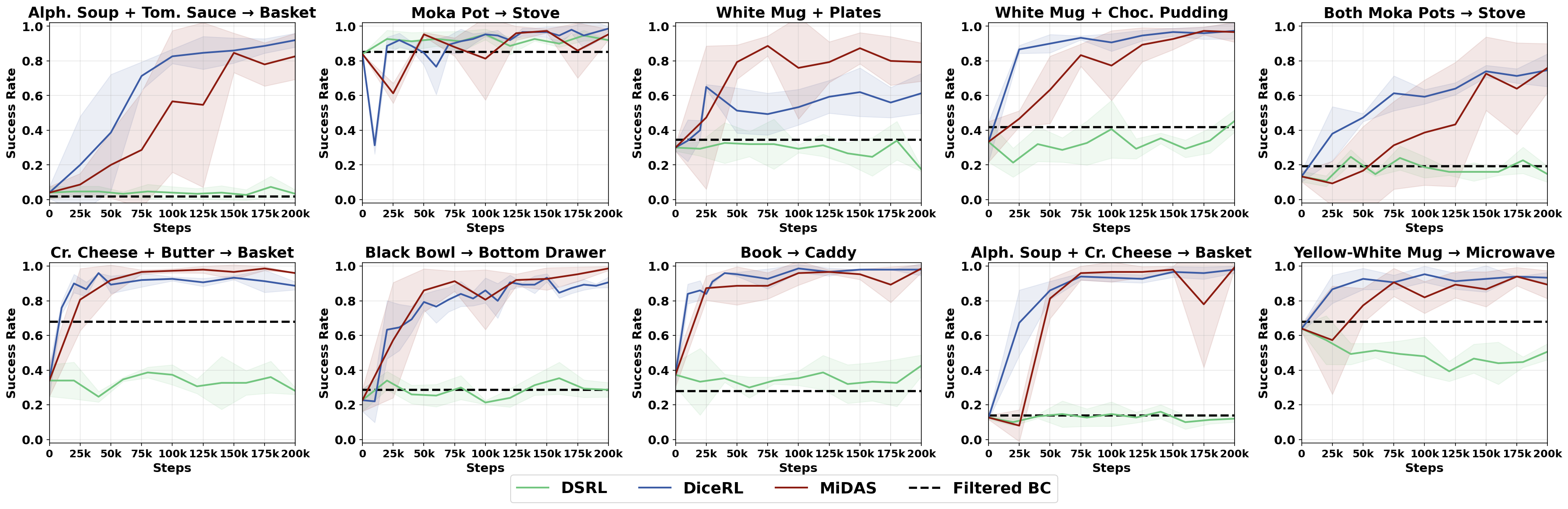}
    \caption{Libero-Long Learning Curves for \midas with baseline comparison. Solid line shows mean across 3 seeds and shaded region denotes standard deviation. Steps on X-axis indicate chunked environment steps (each chunked environment step is equivalent to $C$ actual environment steps)}
    \label{fig:libero_main_table}
\end{figure}

\paragraph{LIBERO-Long.}
LIBERO-Long consists of 10 long-horizon, language-conditioned manipulation tasks performed by a fixed-base Franka Panda arm. Each task requires completing multiple subtasks in sequence (e.g., picking two objects and placing them in a container). The 10 tasks are:
\begin{enumerate}[nosep]
    \item Put both the alphabet soup and the tomato sauce in the basket
    \item Put both the cream cheese box and the butter in the basket
    \item Turn on the stove and put the moka pot on it
    \item Put the black bowl in the bottom drawer of the cabinet and close it
    \item Put the white mug on the left plate and put the yellow and white mug on the right plate
    \item Pick up the book and place it in the back compartment of the caddy
    \item Put the white mug on the plate and put the chocolate pudding to the right of the plate
    \item Put both the alphabet soup and the cream cheese box in the basket
    \item Put both moka pots on the stove
    \item Put the yellow and white mug in the microwave and close it
\end{enumerate}
Each task has 50 human-teleoperated demonstrations available in the benchmark; we use only $K=1$ for adaptation in our main experiments. The environment provides two camera views: an \texttt{agentview} third-person camera and a \texttt{robot0\_eye\_in\_hand} wrist camera, both rendered at $224 \times 224$ pixels. The action space is 7-dimensional: 3D end-effector position, 3D end-effector rotation (axis-angle), and 1D gripper open/close. The episode horizon is 500 timesteps. The learning curves for all the tasks can be visualized in \cref{fig:libero_main_table}.

\begin{table}[t]
\centering
\scriptsize
\renewcommand{\arraystretch}{0.92}
\setlength{\tabcolsep}{2pt}
\caption{\textbf{One demonstration adaptation results measured over 3 seeds.}
Success rates (\%) are mean $\pm$ std.
Note that the zero-shot performance of the pretrained policy is 0\%.}
\label{tab:one_demo_results_libero_with_dicerl}
\resizebox{0.9\columnwidth}{!}{%
\begin{tabular}{l c ccccc}
\toprule
& & \multicolumn{5}{c}{\textbf{1 Demo}} \\
\cmidrule(lr){3-7}
Task & Zero-Shot & BC & DSRL & Filt.\ BC & DiceRL & \textbf{\midas} \\
\midrule
\textit{LIBERO-Long} \\

\quad Alph.\ Soup + Cr.\ Cheese $\to$ Basket
& 0.0
& 12.7\pms{1.2}
& 12.0\pms{2.0}
& 14.0\pms{2.0}
& 98.0\pms{2.0}
& \textbf{99.3\pms{1.2}} \\

\quad Alph.\ Soup + Tom.\ Sauce $\to$ Basket
& 0.0
& 4.0\pms{3.5}
& 3.3\pms{2.3}
& 2.0\pms{0.0}
& \textbf{92.0\pms{4.0}}
& 82.7\pms{13.3} \\

\quad Black Bowl $\to$ Bottom Drawer
& 0.0
& 22.7\pms{6.4}
& 28.7\pms{4.2}
& 28.7\pms{10.3}
& 90.7\pms{3.1}
& \textbf{98.7\pms{1.2}} \\

\quad Book $\to$ Caddy
& 0.0
& 37.3\pms{7.0}
& 42.7\pms{6.1}
& 28.0\pms{2.0}
& 98.0\pms{3.5}
& \textbf{98.7\pms{2.3}} \\

\quad Both Moka Pots $\to$ Stove
& 0.0
& 13.3\pms{3.1}
& 14.7\pms{4.6}
& 19.3\pms{7.0}
& 74.7\pms{9.5}
& \textbf{76.0\pms{14.0}} \\

\quad Cr.\ Cheese + Butter $\to$ Basket
& 0.0
& 34.0\pms{9.2}
& 28.0\pms{2.0}
& 68.0\pms{2.0}
& 88.7\pms{2.3}
& \textbf{96.0\pms{0.0}} \\

\quad Moka Pot $\to$ Stove
& 0.0
& 84.0\pms{0.0}
& 92.0\pms{0.0}
& 85.3\pms{8.3}
& \textbf{98.7\pms{1.2}}
& 95.3\pms{3.1} \\

\quad White Mug + Choc.\ Pudding
& 0.0
& 33.3\pms{11.7}
& 45.3\pms{6.4}
& 42.0\pms{4.0}
& \textbf{97.3\pms{4.6}}
& 96.7\pms{5.8} \\

\quad White Mug + Plates
& 0.0
& 30.0\pms{2.0}
& 17.3\pms{1.2}
& 34.7\pms{6.4}
& 61.3\pms{11.5}
& \textbf{79.3\pms{11.0}} \\

\quad Yellow-White Mug $\to$ Microwave
& 0.0
& 64.0\pms{2.0}
& 50.7\pms{4.6}
& 68.0\pms{7.2}
& \textbf{93.3\pms{3.1}}
& 89.3\pms{8.1} \\

\midrule
\textbf{Average}
& 0.0
& 33.5\pms{1.8}
& 33.5\pms{0.8}
& 39.0\pms{2.4}
& 89.3\pms{1.8}
& \textbf{91.2\pms{4.2}} \\

\bottomrule
\end{tabular}%
}
\end{table}

\textbf{Libero-Long Results Discussion:} We observe that \midas as well as Dice-RL \citep{sun2026priorproefficientskill} enable successful autonomous improvement over the one demonstration warm start. Dice-RL has better sample-efficiency over \midas on 5/10 tasks while under-performing in 1 and equivalent on the remaining 4.

\textbf{DICE-RL.}
We implement DICE-RL~\citep{sun2026priorproefficientskill} following the full set of design choices proposed in the original method, using the frozen VLM embedding as the state representation. DICE-RL is a particularly relevant comparison to \midas as it independently incorporates many of the ingredients we find important for stable minimal-data adaptation: a lightweight residual actor over a frozen base policy, demonstration data retained during online learning through offline data mixing, and value-based policy improvement. Beyond these shared ingredients, we retain DICE-RL's multi-sample expectation training over $K$ policy samples for both the critic bootstrap and actor objective, and filtered TD3+BC regularization~\citep{fujimoto2021minimalistapproachofflinereinforcement} that selectively allows value-improving deviations from the base policy. This provides a strong residual-RL baseline that differs from \midas primarily in how value estimates are used for policy improvement.

\begin{figure}[t]
    \centering
    \includegraphics[width=\linewidth]{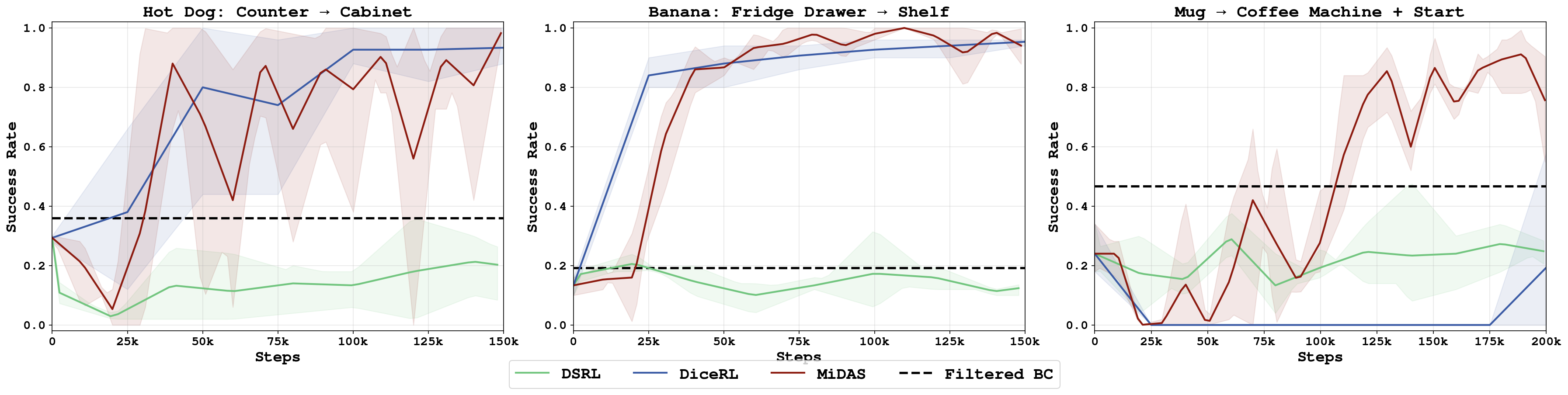}
    \caption{RoboCasa-365 Learning Curves for \midas with baseline comparison. Solid line shows mean across 3 seeds and shaded region denotes standard deviation. Steps on X-axis indicate chunked environment steps (each chunked environment step is equivalent to $C$ actual environment steps)}
    \label{fig:robocasa_main_table}
    \vspace{-1.5em}
\end{figure}

\textbf{RoboCasa-365.}
RoboCasa-365~\citep{nasiriany2026robocasa365largescalesimulationframework} provides long-horizon household manipulation tasks in diverse procedurally generated kitchen scenes. Unlike LIBERO, the robot is a mobile manipulator and tasks involve navigation, object retrieval, and multi-step interaction with kitchen appliances. We evaluate on four tasks:
\begin{enumerate}[nosep]
    \item \textbf{Banana: Fridge Drawer $\to$ Shelf} (\texttt{PickPlaceFridgeDrawerToShelf}): Pick a banana from a fridge drawer and place it on a shelf.
    \item \textbf{Hot Dog: Counter $\to$ Cabinet} (\texttt{PickPlaceCounterToCabinet}): Pick a hot dog from the counter and place it in a cabinet.
    \item \textbf{Mug $\to$ Coffee Machine + Start} (\texttt{PrepareCoffee}): Place a mug under the coffee machine and press the start button.
    \item \textbf{Cup + Bowl $\to$ Dishwasher + Close} (\texttt{LoadDishwasher}): Load a cup and bowl into the dishwasher and close it.
\end{enumerate}
Each task is instantiated in a specific kitchen layout and style that is fixed across training and evaluation. The environment provides two camera views: \texttt{robot0\_agentview\_left} and \texttt{robot0\_eye\_in\_hand}, rendered at $256 \times 256$ pixels. The action space is 7-dimensional for the first three tasks (3D end-effector position, 3D rotation, 1D gripper), with episode horizons varying between 500 and 800 timesteps depending on task complexity (except for \texttt{LoadDishwashwer}). The learning curves for all the tasks can be visualized in \cref{fig:robocasa_main_table}.

\textbf{Long Horizon Manipulation:} Note that the \texttt{LoadDishwasher} task spans a horizon of 1100 timesteps and has an overall success rate of 0\% across all baselines and \midas. This task shows the limitation of our proposed approach and provides motivation for future work to extend \midas to long-horizon, sparse-reward settings.

\paragraph{Shared observation and action details.}
In both benchmarks, the base policy ($\pifive$) receives $224 \times 224$ images after resizing with padding. Actions are temporally chunked: the policy outputs a chunk of $C = 10$ primitive actions that are executed open-loop before replanning (see \cref{app:action-chunking}). The residual learner receives the output of the VLM backbone as the state representation, along with the base policy's proposed action chunk. Evaluation uses 50 rollouts per task from the canonical initial state distribution.

\paragraph{Details on environments.}
For all LIBERO tasks, episodes last at most 500 primitive environment steps.
With action chunk length $H=C=10$ (\cref{tab:parl_hyperparams}), the policy
acts in a chunk-level MDP with at most $500/H=50$ transitions per episode
(\cref{app:action-chunking}). After collecting each trajectory, once
$|\mathcal{B}|>n_{\min}$, we perform one update round per collected transition
using replay minibatches, as specified in Algorithm~\ref{alg:midas}.
Each round comprises
$U_C^{\mathrm{w}}$ critic and $U_A^{\mathrm{w}}$ actor optimizer updates during
warmup ($i<W$), or $U_C$ critic and $U_A$ actor updates thereafter,
and increments $i$ once after these updates.
We use $(U_C^{\mathrm{w}},U_A^{\mathrm{w}})=(8,8)$ and
$(U_C,U_A)=(30,10)$ (\cref{tab:parl_hyperparams}).

\subsection{Libero Pro Perturbation Details}
\label{app:sim_perturb_details}

We use the perturbation protocol from LIBERO-PRO to evaluate generalization under controlled distribution shifts. LIBERO-PRO defines five perturbation axes, of which we use four (excluding task perturbation, which changes the goal itself). We also introduce position perturbations at three magnitudes to test spatial generalization. Not all perturbation types apply to all 10 tasks; LIBERO-PRO defines perturbations only for tasks where a given shift is meaningful. Below we describe each perturbation type along with the specific tasks used.

\paragraph{Observation-level perturbations.}
These change task-irrelevant aspects of the observation while preserving the underlying state and goal.
\begin{itemize}[nosep]
    \item \textbf{Visual (object appearance)} (Fig.~\ref{fig:v2_perturbation_comparison}): Object colors, textures, and sizes are modified in the scene definition (e.g., an \texttt{akita\_black\_bowl} is replaced with a larger variant or a gray \texttt{mokapot} is replaced with a golden one), while the task instruction and goal conditions remain unchanged. We evaluate on 5 tasks where LIBERO-PRO defines color or texture swaps that do not alter the object's shape or grasp affordance:
    \emph{Both Moka Pots $\to$ Stove},
    \emph{Cr.\ Cheese + Butter $\to$ Basket},
    \emph{Moka Pot $\to$ Stove},
    \emph{White Mug + Choc.\ Pudding},
    \emph{White Mug + Plates}.

    \item \textbf{Language} (Fig.~\ref{fig:v2_perturbation_comparison}): The task instruction is replaced with a paraphrase that describes the same goal using different phrasing (e.g., ``put both moka pots on the stove'' $\to$ ``place both moka pots on stove''). Object references and goal logic are preserved. We evaluate on all 10 LIBERO-Long tasks, as language paraphrases are defined for every task.
\end{itemize}

\paragraph{State-level perturbations.}
These change dynamically relevant quantities that may require different actions. LIBERO-PRO applies a single object-replacement perturbation per task, but the nature of the replacement varies: for some tasks the replacement changes only the object's shape while preserving its category, for others it substitutes a different object category entirely, and for others the two scene objects swap positions. We group results by the type of change each replacement induces.
\begin{itemize}[nosep]
    \item \textbf{Object swap} (Fig.~\ref{fig:v2_perturbation_comparison}): The initial positions of two objects are exchanged by swapping their spawn-region assignments in the scene definition. The policy must execute the task with objects in different spatial locations. We evaluate on the same 5 tasks as the visual perturbation:
    \emph{Both Moka Pots $\to$ Stove},
    \emph{Cr.\ Cheese + Butter $\to$ Basket},
    \emph{Moka Pot $\to$ Stove},
    \emph{White Mug + Choc.\ Pudding},
    \emph{White Mug + Plates}.

    \item \textbf{Shape change} (Fig.~\ref{fig:v2_perturbation_comparison}): Target object meshes are replaced with geometrically different variants of the same category (e.g., a taller mug or wider bowl). If the new shape preserves similar grasp affordances, the required manipulation strategy remains the same; otherwise the policy must adapt its contact behavior. We evaluate on 4 tasks where the LIBERO-PRO replacement alters the object geometry:
    \emph{Black Bowl $\to$ Bottom Drawer},
    \emph{Yellow-White Mug $\to$ Microwave},
    \emph{Alph.\ Soup + Cr.\ Cheese $\to$ Basket},
    \emph{Alph.\ Soup + Tom.\ Sauce $\to$ Basket}.

    \item \textbf{Object change} (Fig.~\ref{fig:v2_perturbation_comparison}): The target object is replaced with a different object category entirely, changing both visual appearance and manipulation affordances. We evaluate on 1 task where the replacement introduces a categorically different object: \emph{Book $\to$ Caddy}.
\end{itemize}

\paragraph{Position perturbation.}
We evaluate position generalization by displacing target objects from their canonical initial positions at three levels:
\begin{itemize}[nosep]
    \item \textbf{Low:} 5\,cm XY displacement, 0.1 radians yaw perturbation
    \item \textbf{Medium:} 10\,cm XY displacement, 0.15 radians yaw perturbation
    \item \textbf{High:} 20\,cm XY displacement, 0.25 radians yaw perturbation
\end{itemize}
XY offsets are sampled uniformly within a disk of the specified radius. After applying perturbations, objects are settled under physics simulation and validated: the perturbed object must remain visible in the camera frame, stay within 2\,cm of its settled z-height (no falling off surfaces), not collide with other scene objects, and maintain an upright orientation (local up-axis within $18^\circ$ of the world vertical). Invalid samples are re-drawn for up to 50 attempts per episode.

We evaluate position perturbation on 3 tasks:
\emph{Alph.\ Soup + Tom.\ Sauce $\to$ Basket},
\emph{Black Bowl $\to$ Bottom Drawer},
\emph{Both Moka Pots $\to$ Stove}.
Perturbations are applied to the task-relevant target objects in each case (e.g., both moka pots, or the black bowl).

\paragraph{Evaluation protocol.}
All perturbation evaluations use 50 rollouts per task. For LIBERO-PRO perturbations (visual, language, swap, shape, object change), we evaluate from the perturbed initial configurations using the standard episode horizon. For position perturbations, we evaluate at each magnitude separately. Results are averaged across 3 seeds. Per-task breakdowns are provided in the tables below.

\paragraph{Visual perturbation results.}
Table~\ref{tab:visual_perturb} reports per-task success rates under visual (object appearance) perturbation.

\begin{table}[h]
\centering
\begin{tabular}{l cc cc}
\toprule
 & \multicolumn{2}{c}{$\pibase^1$} & \multicolumn{2}{c}{$\pi_{\mathrm{RL}}^1$} \\
\cmidrule(lr){2-3} \cmidrule(lr){4-5}
Task & ID & OOD & ID & OOD \\
\midrule
Both Moka Pots $\to$ Stove & 13.3\pms{3.1} & 6.0\pms{2.0} & 76.0\pms{14.0} & 58.7\pms{33.5} \\
Cr.\ Cheese + Butter $\to$ Basket & 34.0\pms{9.2} & 38.0\pms{9.2} & 96.0\pms{0.0} & 98.7\pms{2.3} \\
Moka Pot $\to$ Stove & 84.0\pms{0.0} & 58.7\pms{5.0} & 95.3\pms{3.1} & 93.3\pms{4.2} \\
White Mug + Choc.\ Pudding & 33.3\pms{11.7} & 30.0\pms{10.4} & 96.7\pms{5.8} & 80.0\pms{11.1} \\
White Mug + Plates & 30.0\pms{2.0} & 24.0\pms{2.0} & 79.3\pms{11.0} & 79.3\pms{13.6} \\
\midrule
\textbf{Average} & 38.9\pms{3.6} & 31.3\pms{4.4} & 88.7\pms{5.9} & 82.0\pms{8.0} \\
\bottomrule
\end{tabular}
\caption{Per-task success rate (\%) under visual (object appearance) perturbation. OOD replaces object colors, textures, or sizes while preserving shape and task goal.}
\label{tab:visual_perturb}
\end{table}

\paragraph{Language perturbation results.}
Table~\ref{tab:lang_perturb} reports per-task success rates under language perturbation.

\begin{table}[h]
\centering
\begin{tabular}{l cc cc}
\toprule
 & \multicolumn{2}{c}{$\pibase^1$} & \multicolumn{2}{c}{$\pi_{\mathrm{RL}}^1$} \\
\cmidrule(lr){2-3} \cmidrule(lr){4-5}
Task & ID & OOD & ID & OOD \\
\midrule
Alph.\ Soup + Cr.\ Cheese $\to$ Basket & 12.7\pms{1.2} & 14.0\pms{2.0} & 99.3\pms{1.2} & 98.0\pms{2.0} \\
Alph.\ Soup + Tom.\ Sauce $\to$ Basket & 4.0\pms{3.5} & 4.7\pms{3.1} & 82.7\pms{13.3} & 78.7\pms{11.0} \\
Black Bowl $\to$ Bottom Drawer & 22.7\pms{6.4} & 23.3\pms{9.2} & 98.7\pms{1.2} & 96.7\pms{5.8} \\
Book $\to$ Caddy & 37.3\pms{7.0} & 33.3\pms{8.3} & 98.7\pms{2.3} & 96.7\pms{4.2} \\
Both Moka Pots $\to$ Stove & 13.3\pms{3.1} & 15.3\pms{2.3} & 76.0\pms{14.0} & 79.3\pms{13.0} \\
Cr.\ Cheese + Butter $\to$ Basket & 34.0\pms{9.2} & 40.0\pms{14.0} & 96.0\pms{0.0} & 96.0\pms{2.0} \\
Moka Pot $\to$ Stove & 84.0\pms{0.0} & 88.7\pms{5.8} & 95.3\pms{3.1} & 97.3\pms{2.3} \\
White Mug + Choc.\ Pudding & 33.3\pms{11.7} & 38.7\pms{13.3} & 96.7\pms{5.8} & 91.3\pms{8.1} \\
White Mug + Plates & 30.0\pms{2.0} & 33.3\pms{5.8} & 79.3\pms{11.0} & 78.0\pms{2.0} \\
Yellow-White Mug $\to$ Microwave & 64.0\pms{2.0} & 56.0\pms{6.9} & 89.3\pms{8.1} & 90.0\pms{7.2} \\
\midrule
\textbf{Average} & 33.5\pms{1.8} & 34.7\pms{4.8} & 91.2\pms{4.2} & 90.2\pms{3.3} \\
\bottomrule
\end{tabular}
\caption{Per-task success rate (\%) under language perturbation. OOD replaces the task instruction with a paraphrase.}
\label{tab:lang_perturb}
\end{table}

\subsection{Libero Pro State Shift Perturbation Experiments}
\label{app:sim_perturb_exps}

Here we provide further analysis of the state-shift generalization results from Section~\ref{sec:generalization}.

\paragraph{Object swap.}
When two objects exchange positions, the policy must approach the target from a different direction or reorder its subtask sequence. On the 5 evaluated tasks (\emph{Both Moka Pots $\to$ Stove}, \emph{Cr.\ Cheese + Butter $\to$ Basket}, \emph{Moka Pot $\to$ Stove}, \emph{White Mug + Choc.\ Pudding}, \emph{White Mug + Plates}), swapping object positions changes the optimal grasp order and approach trajectory. Both $\pibase^1$ and $\pi_{\mathrm{RL}}^1$ achieve 0\% success across all 5 tasks under swap perturbation (Table~\ref{tab:swap_perturb_appendix}), despite strong ID performance. This indicates that swapping object positions produces configurations far enough from the training distribution that neither the BC warm start nor residual RL corrections can recover.

\begin{table}[h]
\centering
\begin{tabular}{l cc cc}
\toprule
 & \multicolumn{2}{c}{$\pibase^1$} & \multicolumn{2}{c}{$\pi_{\mathrm{RL}}^1$} \\
\cmidrule(lr){2-3} \cmidrule(lr){4-5}
Task & ID & OOD & ID & OOD \\
\midrule
Both Moka Pots $\to$ Stove & 13.3\pms{3.1} & 0.0\pms{0.0} & 76.0\pms{14.0} & 0.0\pms{0.0} \\
Cr.\ Cheese + Butter $\to$ Basket & 34.0\pms{9.2} & 0.0\pms{0.0} & 96.0\pms{0.0} & 0.0\pms{0.0} \\
Moka Pot $\to$ Stove & 84.0\pms{0.0} & 0.0\pms{0.0} & 95.3\pms{3.1} & 0.0\pms{0.0} \\
White Mug + Choc.\ Pudding & 33.3\pms{11.7} & 0.0\pms{0.0} & 96.7\pms{5.8} & 0.0\pms{0.0} \\
White Mug + Plates & 30.0\pms{2.0} & 0.0\pms{0.0} & 79.3\pms{11.0} & 0.0\pms{0.0} \\
\midrule
\textbf{Average} & 38.9\pms{3.6} & 0.0\pms{0.0} & 88.7\pms{5.9} & 0.0\pms{0.0} \\
\bottomrule
\end{tabular}
\caption{Per-task success rate (\%) under object swap perturbation. OOD exchanges the initial positions of two scene objects.}
\label{tab:swap_perturb_appendix}
\end{table}

\paragraph{Shape change.}
Shape perturbations test whether learned grasps transfer across object geometries. On the 4 evaluated tasks (\emph{Black Bowl $\to$ Bottom Drawer}, \emph{Yellow-White Mug $\to$ Microwave}, \emph{Alph.\ Soup + Cr.\ Cheese $\to$ Basket}, \emph{Alph.\ Soup + Tom.\ Sauce $\to$ Basket}), the replacement preserves the object category but alters its geometry. Table~\ref{tab:shape_perturb_appendix} shows that $\pi_{\mathrm{RL}}^1$ retains strong performance on tasks where the new shape preserves similar grasp affordances (e.g., Black Bowl at 90.7\%), but drops sharply when the shape change alters the required contact geometry (e.g., Alph.\ Soup tasks at $<$3\%).

\begin{table}[h]
\centering
\begin{tabular}{l cc cc}
\toprule
 & \multicolumn{2}{c}{$\pibase^1$} & \multicolumn{2}{c}{$\pi_{\mathrm{RL}}^1$} \\
\cmidrule(lr){2-3} \cmidrule(lr){4-5}
Task & ID & OOD & ID & OOD \\
\midrule
Black Bowl $\to$ Bottom Drawer & 22.7\pms{6.4} & 20.7\pms{6.4} & 98.7\pms{1.2} & 90.7\pms{8.1} \\
Yellow-White Mug $\to$ Microwave & 64.0\pms{2.0} & 4.7\pms{3.1} & 89.3\pms{8.1} & 20.7\pms{13.3} \\
Alph.\ Soup + Cr.\ Cheese $\to$ Basket & 12.7\pms{1.2} & 0.0\pms{0.0} & 99.3\pms{1.2} & 2.7\pms{4.6} \\
Alph.\ Soup + Tom.\ Sauce $\to$ Basket & 4.0\pms{3.5} & 0.0\pms{0.0} & 82.7\pms{13.3} & 2.0\pms{3.5} \\
\midrule
\textbf{Average} & 25.8\pms{0.8} & 6.3\pms{1.0} & 92.5\pms{5.4} & 29.0\pms{5.8} \\
\bottomrule
\end{tabular}
\caption{Per-task success rate (\%) under shape change perturbation. OOD replaces the target object with a geometrically different variant of the same category.}
\label{tab:shape_perturb_appendix}
\end{table}

\paragraph{Object change.}
Replacing the target with a different object category represents the most severe state shift. We evaluate this on \emph{Book $\to$ Caddy}, where the book is replaced with a categorically different object. Table~\ref{tab:obj_change_perturb_appendix} shows that both $\pibase^1$ and $\pi_{\mathrm{RL}}^1$ drop to near 0\% under this shift despite strong ID performance. The new object differs in size, shape, weight, and affordance, requiring manipulation strategies outside the behavioral support of $\pi_{\mathrm{RL}}^1$. This category of shift cannot be addressed by online RL alone and requires additional demonstrations or interaction with the new object.

\begin{table}[h]
\centering
\small
\begin{tabular}{l cc cc}
\toprule
 & \multicolumn{2}{c}{$\pibase^1$} & \multicolumn{2}{c}{$\pi_{\mathrm{RL}}^1$} \\
\cmidrule(lr){2-3} \cmidrule(lr){4-5}
Task & ID & OOD & ID & OOD \\
\midrule
Book $\to$ Caddy & 37.3\pms{7.0} & 0.0\pms{0.0} & 98.7\pms{2.3} & 2.0\pms{2.0} \\
\bottomrule
\end{tabular}
\caption{Per-task success rate (\%) under object change perturbation. OOD replaces the target with a categorically different object.}
\label{tab:obj_change_perturb_appendix}
\end{table}

\paragraph{Position perturbation.}
Tables~\ref{tab:pos_perturb_low},\ref{tab:pos_perturb_medium},\ref{tab:pos_perturb_high} reports per-task success rates under position perturbation at three displacement levels. We compare $\pibase^1$, $\pi_{\mathrm{RL}}^1$, and $\pibase^{50}$ (behavior-cloned on all 50 demonstrations). $\pi_{\mathrm{RL}}^1$ improves over $\pibase^1$ at all levels but falls short of $\pibase^{50}$, especially at larger displacements where the gap in state coverage becomes the binding constraint.

\begin{table}[h]
\centering
\small
\begin{tabular}{l ccc}
\toprule
Task & $\pibase^1$ & $\pi_{\mathrm{RL}}^1$ & $\pibase^{50}$ \\
\midrule
Alph.\ Soup + Tom.\ Sauce
& 4.0\pms{2.0} & 52.7\pms{5.8} & 94.7\pms{5.8} \\

Black Bowl $\to$ Drawer
& 24.0\pms{7.2} & 88.7\pms{5.0} & 59.3\pms{10.1} \\

Both Moka Pots $\to$ Stove
& 3.3\pms{3.1} & 49.3\pms{20.1} & 84.0\pms{7.2} \\
\midrule
\textbf{Average}
& 10.4\pms{2.5} & 63.6\pms{7.3} & 79.3\pms{6.4} \\
\bottomrule
\end{tabular}
\caption{Per-task success rate (\%) under low position perturbation of 5\,cm.}
\label{tab:pos_perturb_low}
\end{table}

\begin{table}[h]
\centering
\small
\begin{tabular}{l ccc}
\toprule
Task & $\pibase^1$ & $\pi_{\mathrm{RL}}^1$ & $\pibase^{50}$ \\
\midrule
Alph.\ Soup + Tom.\ Sauce
& 1.3\pms{2.3} & 14.0\pms{2.0} & 68.7\pms{3.1} \\

Black Bowl $\to$ Drawer
& 16.7\pms{6.4} & 70.0\pms{3.5} & 45.3\pms{2.3} \\

Both Moka Pots $\to$ Stove
& 1.3\pms{1.2} & 22.7\pms{8.3} & 76.0\pms{7.2} \\
\midrule
\textbf{Average}
& 6.4\pms{3.1} & 35.6\pms{1.5} & 63.3\pms{1.2} \\
\bottomrule
\end{tabular}
\caption{Per-task success rate (\%) under medium position perturbation of 10\,cm.}
\label{tab:pos_perturb_medium}
\end{table}

\begin{table}[h]
\centering
\small
\begin{tabular}{l ccc}
\toprule
Task & $\pibase^1$ & $\pi_{\mathrm{RL}}^1$ & $\pibase^{50}$ \\
\midrule
Alph.\ Soup + Tom.\ Sauce
& 0.0\pms{0.0} & 2.7\pms{3.1} & 25.3\pms{4.2} \\

Black Bowl $\to$ Drawer
& 6.7\pms{2.3} & 37.3\pms{2.3} & 23.3\pms{8.1} \\

Both Moka Pots $\to$ Stove
& 0.0\pms{0.0} & 2.7\pms{3.1} & 25.3\pms{9.5} \\
\midrule
\textbf{Average}
& 2.2\pms{0.8} & 14.2\pms{0.4} & 24.7\pms{4.4} \\
\bottomrule
\end{tabular}
\caption{Per-task success rate (\%) under high position perturbation of 20\,cm.}
\label{tab:pos_perturb_high}
\end{table}

\section{Implementation Details}
\label{app:impl_details}

\subsection{MiDaS}
\label{app:parl_details}

\textbf{State representation.}
The state input to both the residual actor and critic is derived from the frozen VLA backbone of $\pibase^K$ along with the proprioceptive state.
At each timestep, we extract the VLM token embeddings and mean-pool across all token positions to produce a 2048-dimensional feature vector.
This vector is then projected through a bottleneck layer (linear projection, layer normalization, and tanh activation) to yield a 200-dimensional latent state which is then concatenated with the proprioceptive state to get the state representation $\bs_t$.

\textbf{Actor inputs and architecture.}
The residual actor receives the state representation as the bottlenecked VLM embedding and proprioceptive state, concatenated with the flattened base action chunk $\ba_t^{\mathrm{base}} \in \R^{C \cdot d_a}$ proposed by $\pibase^K$, where $C{=}10$ is the chunk length and $d_a{=}7$ is the per-step action dimension.
As described in \cref{sec:midas}, the actor directly outputs the full executed action chunk $\ba_t^{\mathrm{exec}}$ rather than an additive residual correction.
The actor is parameterized as a 3-layer MLP with 512 hidden units per layer and ReLU activations.
The output distribution is a tanh-squashed diagonal Gaussian with learned mean and log-standard-deviation heads.

\textbf{Critic inputs and architecture.}
The critic receives only the bottlenecked VLM embedding and proprioceptive state as its state input; the base action chunk is excluded.
Given a candidate executed action $\ba_t^{\mathrm{exec}}$, each Q-network in the ensemble evaluates $Q_{\psi_j}(\bs_t, \ba_t^{\mathrm{exec}})$.
We use an ensemble of 10 Q-networks, each parameterized as a 3-layer MLP with 512 hidden units and ReLU activations.
The ensemble aggregates predictions via mean reduction during target computation.

\textbf{Warmup Details.} We collect warmup data by rolling out the base policy with an action chunk length of $C=10$. We count collection steps at the chunk level: each step corresponds to executing one action chunk and collecting one chunked transition, spanning $10$ primitive environment steps. We collect a total of $10{,}000$ chunked transitions, corresponding to $100{,}000$ primitive environment steps when all chunks are executed in full. For LIBERO, the maximum episode horizon is $500$ primitive environment steps, or $50$ chunks. Thus, this collection budget is equivalent to approximately $200$ full-length episodes; collecting the same number of transitions requires more episodes if trajectories terminate early. For Robocasa, except for \texttt{Cup + Bowl $\to$ Dishwasher + Close} and \texttt{Mug $\to$ Coffee Machine + Start}, we use a horizon of $500$. The given two tasks use $1100$ and $800$ respectively. During warmup, we perform eight actor optimizer updates and eight critic optimizer updates per collected chunked transition. The update-to-data (UTD) ratio is therefore $8$ for each network when measured per chunked transition. Normalized by primitive environment steps, this corresponds to $8/C=0.8$ updates per environment step for each network.

\textbf{Policy update.}
At each actor update, a single base action $\ba_t^{\mathrm{base}}$ is sampled from the frozen $\pibase^K$ (in practice we use the cached base action from the replay buffer).
The residual actor then proposes $N{=}16$ candidate actions conditioned on $\bs_t$ and $\ba_t^{\mathrm{base}}$, and also includes the base action as a valid proposal.
The top $M{=}8$ actions ranked by the critic ensemble are selected as elites.
These elite actions undergo 30 steps of gradient ascent on the Q-ensemble with step size $10^{-3}$.
The actor is then updated via a distillation loss that regresses toward the refined elite action.

\textbf{Optimization.}
Both actor and critic use the Adam optimizer with global gradient norm clipping at 1.0.
The actor learning rate is $10^{-4}$ and the critic learning rate is $3 \times 10^{-4}$.
For each chunked environment step (equivalent to $C$ actual environment steps), we perform 30 critic updates and 10 actor updates.
Training begins with a BC warmup phase as described in the warmup details above.

\textbf{Key hyperparameters.}
\Cref{tab:parl_hyperparams} summarizes the hyperparameters used for all PARL training runs.

\begin{table}[h]
\centering
\small
\caption{PARL hyperparameters used for all training runs.}
\label{tab:parl_hyperparams}
\begin{tabular}{ll}
\toprule
\textbf{Hyperparameter} & \textbf{Value} \\
\midrule
Batch size ($B$) & 64 \\
Discount factor ($\gamma$) & 0.999 \\
Target network update rate ($\tau$) & 0.05 \\
Actor learning rate ($\alpha_\pi$) & $10^{-4}$ \\
Critic learning rate & $3 \times 10^{-4}$ \\
Hidden dimensions & $(512, 512, 512)$ \\
VLM embedding bottleneck dim & 200 \\
Number of Q-networks ($J$) & 10 \\
Critic ensemble reduction & mean \\
PA-RL candidate samples ($N$) & 16 \\
PA-RL elite samples ($M$) & 8 \\
PA-RL gradient steps ($G$) & 30 \\
PA-RL step size ($\eta$) & $10^{-3}$ \\
Critic updates per chunked step ($U_C$) & 30 \\
Actor updates per chunked step ($U_A$) & 10 \\
Warmup collection steps (chunked) & 10{,}000 \\
BC warmup critic/actor updates ($U_C^{\mathrm{w}}, U_A^{\mathrm{w}}$) & 8 \\
Success buffer ratio ($\rho_{\mathrm{succ}}$) & 0.20 \\
Success buffer minimum size & 20 \\
Gradient norm clip & 1.0 \\
Optimizer & Adam \\
Action chunk length ($C$, $H$) & 10 \\
\bottomrule
\end{tabular}
\end{table}

\subsection{BC Implementation}
\label{app:bc_details}
For behavior cloning, we use a single demonstration out of the available 50 for LIBERO-Long and one demonstration for the particular scene and layout configuration for each RoboCasa task. We then fine-tune $\pifive$ using the official OpenPI codebase on this demonstration for 16{,}000 gradient steps. \cref{tab:libero10_demos_full} shows the performance on of behavior cloning on varying the number of demos $K=\{1,3,5\}$.

\textbf{Trainable parameters.}
We apply LoRA adapters to the PaliGemma 2B vision-language-model (VLM) backbone, inserting low-rank updates into both the attention and feed-forward layers with rank 32 and scaling factor $\alpha{=}32$.
The 300M-parameter action head (Gemma 300M) and the SigLIP vision encoder are fully fine-tuned.
All other backbone weights remain frozen. This selective finetuning keeps the majority of the VLM fixed while allowing the action head and vision encoder to specialize to the target task.

\textbf{Loss and optimization.}
The model is trained with the flow-matching objective native to the $\pifive$ architecture, predicting action chunks of horizon $C{=}10$.
We use AdamW with $\beta_1{=}0.9$, $\beta_2{=}0.95$, and negligible weight decay ($10^{-10}$), with global gradient norm clipping at 1.0.
The learning rate follows a cosine decay schedule: peak learning rate $2.5 \times 10^{-5}$, 50 warmup steps, decaying to $2.5 \times 10^{-6}$ over 100{,}000 steps. Training is stopped at 16{,}000 gradient steps. The batch size is 32.

\textbf{Key hyperparameters.}
\Cref{tab:bc_hyperparams} lists the hyperparameters used for all Stage I behavior cloning runs.

\begin{table}[h]
\centering
\small
\caption{Stage I behavior cloning hyperparameters.}
\label{tab:bc_hyperparams}
\begin{tabular}{ll}
\toprule
\textbf{Hyperparameter} & \textbf{Value} \\
\midrule
Number of demonstrations ($K$) & 1 \\
Training steps & 16{,}000 \\
Batch size & 32 \\
Optimizer & AdamW \\
Peak learning rate & $2.5 \times 10^{-5}$ \\
LR schedule & Cosine decay \\
LR warmup steps & 50 \\
Minimum learning rate & $2.5 \times 10^{-6}$ \\
$\beta_1, \beta_2$ & 0.9, 0.95 \\
Gradient norm clip & 1.0 \\
LoRA rank & 32 \\
LoRA $\alpha$ & 32 \\
LoRA targets & Attention + FFN \\
Action expert & Fully fine-tuned \\
Vision encoder (SigLIP) & Fully fine-tuned \\
Action horizon ($C$) & 10 \\
Loss & Flow matching \\
\bottomrule
\end{tabular}
\end{table}

\begin{table}[h]
\centering
\begin{tabular}{l ccc}
\toprule
 & \multicolumn{3}{c}{\textbf{BC Success Rate (\%)}} \\
\cmidrule(lr){2-4}
Task & $K{=}1$ & $K{=}3$ & $K{=}5$ \\
\midrule
\textit{LIBERO-Long} \\
\quad Alphabet Soup + Cream Cheese $\to$ Basket & 12.7 & 39.3 & \textbf{82.7} \\
\quad Alphabet Soup + Tomato Sauce $\to$ Basket & 4.0 & 32.7 & \textbf{74.7} \\
\quad Black Bowl $\to$ Bottom Drawer & 22.7 & 90.0 & \textbf{94.0} \\
\quad Book $\to$ Caddy & 37.3 & 26.7 & \textbf{89.3} \\
\quad Both Moka Pots $\to$ Stove & 13.3 & \textbf{51.3} & 35.3 \\
\quad Cream Cheese + Butter $\to$ Basket & 34.0 & \textbf{90.7} & 84.7 \\
\quad Moka Pot $\to$ Stove & 84.0 & 86.0 & \textbf{94.7} \\
\quad White Mug + Chocolate Pudding & 33.3 & 26.7 & \textbf{66.0} \\
\quad White Mug + Plates & 30.0 & 48.7 & \textbf{74.0} \\
\quad Yellow-White Mug $\to$ Microwave & 64.0 & 70.7 & \textbf{86.0} \\
\midrule
\textbf{Average} & 33.5 & 56.3 & \textbf{78.1} \\
\bottomrule
\end{tabular}
\caption{\textbf{Per-task BC success on LIBERO-Long.} Bold denotes the best result per task across $K \in \{1,3,5\}$ demonstrations.}
\label{tab:libero10_demos_full}
\end{table}

\subsection{Success Buffer}
\label{app:success_buffer}

The success buffer $\mathcal{B}_{\mathrm{succ}}$ is a separate replay buffer that stores only transitions from successful episodes. Its purpose is to prevent actor and critic updates from being dominated by failed rollouts when the policy's success rate is low (2--5\% early in training).

\textbf{Initialization.}
Both the main replay buffer $\mathcal{B}$ and the success buffer $\mathcal{B}_{\mathrm{succ}}$ are initialized with all transitions from the $K$ demonstrations, since demonstration trajectories are successful by construction. The success buffer capacity is set to $\max(10{,}000,\; 0.5 \times |\mathcal{B}|_{\mathrm{max}})$.

\textbf{Population during training.}
After each collected trajectory, all transitions are inserted into $\mathcal{B}$. If the episode was successful (sparse reward received), the same transitions are additionally inserted into $\mathcal{B}_{\mathrm{succ}}$.

\textbf{Sampling.}
Actor and critic update minibatches are sampled from a mixture of the two buffers:
\begin{align}
\label{eqn:success_buffer}
    \mathcal{B}_{\mathrm{actor/critic}} \sim (1 - \rho_{\mathrm{succ}}) \cdot \mathcal{B} + \rho_{\mathrm{succ}} \cdot \mathcal{B}_{\mathrm{succ}},
\end{align}
where $\rho_{\mathrm{succ}} = 0.2$ is the success buffer ratio. Concretely, each actor batch of size 64 contains 51 transitions sampled uniformly from $\mathcal{B}$ and 13 transitions sampled uniformly from $\mathcal{B}_{\mathrm{succ}}$. This oversampling biases the actor toward states where successful behavior has been observed, steering the PA-RL elite selection and gradient ascent toward reward-bearing regions.

\section{Ablations}
\label{app:ablations}

\begin{figure}[t]
    \centering
    \includegraphics[width=0.6\linewidth]{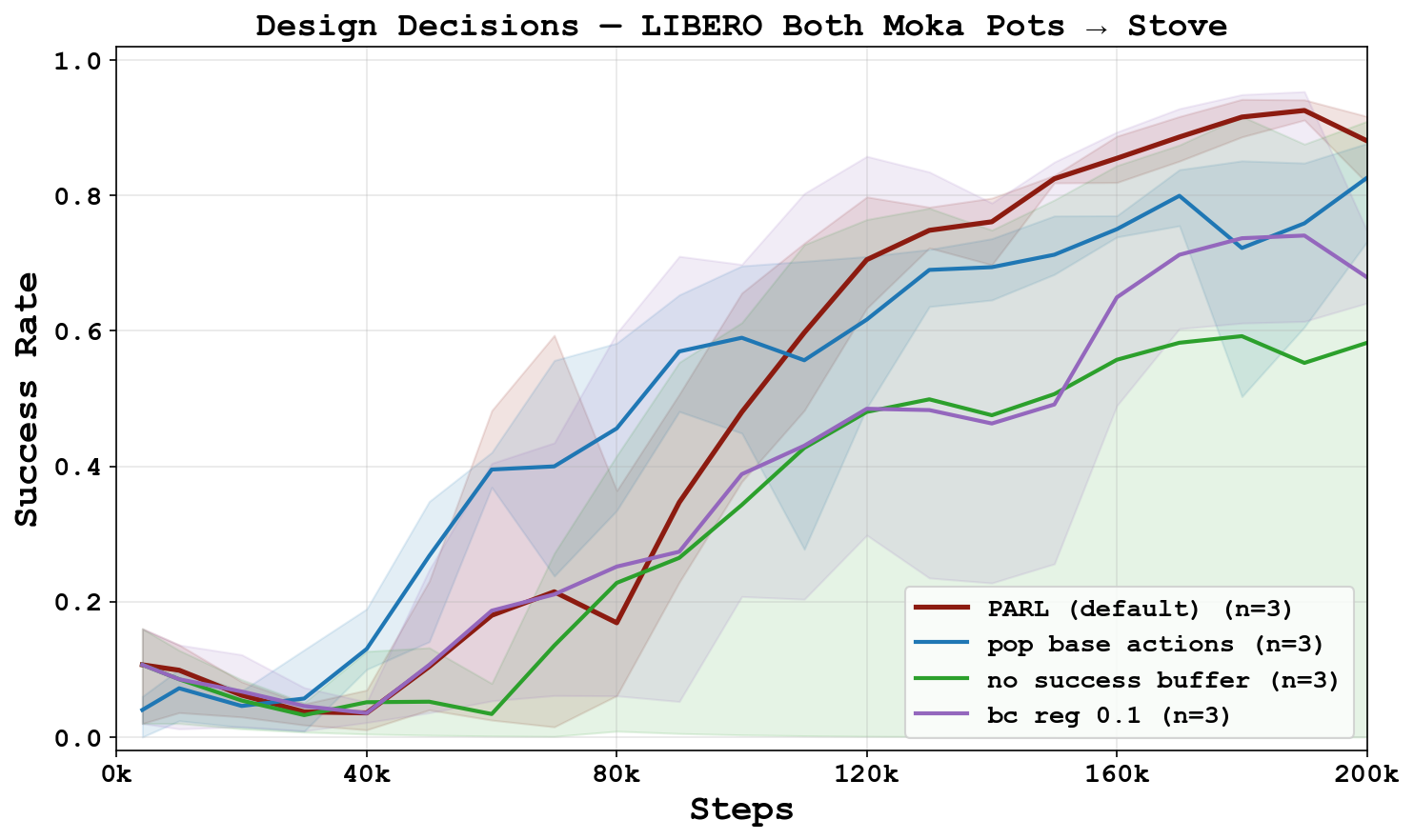}
    \caption{\midas ablations on both-mokapots task}
    \label{fig:mokapots_ablation}
\end{figure}

\confpar{Design-decision ablations.}
\cref{fig:mokapots_ablation} ablates three implementation choices in MIDAS on the \texttt{Both Moka Pots -> Stove} task. Each row changes a single knob while keeping the same PA-RL objective, VLM state representation, 10k-step BC warmup, and one demonstration initialization. First, removing the base-policy action from the residual actor nearly matches the default recipe. This suggests that the action head is most important for producing task-relevant warmup experience and distilling this information into the residual actor during the warmup; once this experience is available, online value learning from pretrained VLA features can recover much of the final performance without using the base action as an explicit actor input. Second, disabling the success buffer tests whether sparse successful transitions must be oversampled during actor updates, rather than being learned from uniformly mixed online replay. Third, adding a persistent success-only BC penalty tests whether the residual should remain behaviorally anchored after warmup, or whether continued RL updates should be allowed to move away from the demonstrated action distribution. 

\section{Curriculum Enables Stronger Position Generalization}
\label{app:curriculum_gen}

\Cref{tab:pos_perturb_low}, \cref{tab:pos_perturb_medium}, \cref{tab:pos_perturb_high} and \cref{fig:v2_pos_gen} show that $\pi_{\mathrm{RL}}^1$ generalizes well to small position perturbations: at 5\,cm displacement, average success reaches 63.6\%, compared to 10.4\% for $\pibase^1$. Performance drops sharply at larger displacements, falling to 35.6\% at 10\,cm and 14.2\% at 20\,cm. By contrast, $\pibase^{50}$ maintains 79.3\%, 63.3\%, and 24.7\% at the same levels, indicating that broader training coverage sustains robustness where $\pi_{\mathrm{RL}}^1$ does not.

A key observation is that larger position perturbations in our setting can be decomposed incrementally: a 10\,cm displacement is reachable through a sequence of smaller shifts from the nominal configuration. If the policy can reliably solve the task under 5\,cm perturbation, it should be able to extend to 10\,cm once it has practiced under the smaller shift. This motivates a curriculum over the reset distribution: rather than training under the full perturbation range from the outset, we progressively widen the displacement radius as the policy masters each level.

\textbf{Curriculum procedure.} Starting from the trained $\pi_{\mathrm{RL}}^1$ checkpoint (after Stage II of \midas), we continue online RL under position-perturbed resets. The curriculum proceeds through three stages with increasing perturbation radii: 5\,cm, 10\,cm, and 20\,cm. At each stage, the target objects are displaced uniformly at random within the specified radius from their default positions. The policy advances to the next stage once its evaluation success rate on the current perturbation level exceeds a threshold of 0.6. On the final stage, training continues until the step budget is exhausted. All other hyperparameters (critic/actor updates, learning rates, success buffer) remain identical to the standard \midas Stage II configuration (\cref{tab:parl_hyperparams}). No additional demonstrations are provided; the curriculum relies entirely on autonomous interaction.

\textbf{Task selection.} We run and evaluate the curriculum on the three tasks used for position perturbation evaluation: \emph{Alphabet Soup + Tomato Sauce $\to$ Basket}, \emph{Both Moka Pots $\to$ Stove}, and \emph{Black Bowl $\to$ Bottom Drawer}. For \emph{Black Bowl}, $\pi_{\mathrm{RL}}^1$ already achieves 88.7\% at 5\,cm and 70.0\% at 10\,cm without curriculum (\cref{tab:pos_perturb_low}, \cref{tab:pos_perturb_medium}), both exceeding the 0.6 advancement threshold. We therefore exclude it from the curriculum experiment, as it already demonstrates strong position generalization under the standard \midas recipe, and use it only for evaluation.

\textbf{Results.} \Cref{tab:curriculum_results} reports per-task success rates after curriculum training, compared against the non-curriculum $\pi_{\mathrm{RL}}^1$ and $\pibase^{50}$ baselines from \cref{tab:pos_perturb_low},\cref{tab:pos_perturb_medium},\cref{tab:pos_perturb_high}. On \emph{Both Moka Pots}, curriculum raises 5\,cm success from 49.3\% to 70.0\% and 10\,cm success from 22.7\% to 75.3\%, nearly matching $\pibase^{50}$ (76.0\%) at the medium perturbation level. On \emph{Alphabet Soup + Tomato Sauce}, 5\,cm success improves from 52.7\% to 68.7\% and 10\,cm success from 14.0\% to 28.7\%. The gains are consistent across seeds, with the largest improvements at the 10\,cm level where the non-curriculum policy struggles most.

\begin{table}[h]
\centering
\begin{tabular}{l cc cc cc}
\toprule
 & \multicolumn{2}{c}{$\pi_{\mathrm{RL}}^1$} & \multicolumn{2}{c}{$\pi_{\mathrm{RL}}^1$ + Curriculum} & \multicolumn{2}{c}{$\pibase^{50}$} \\
\cmidrule(lr){2-3} \cmidrule(lr){4-5} \cmidrule(lr){6-7}
Task & 5\,cm & 10\,cm & 5\,cm & 10\,cm & 5\,cm & 10\,cm \\
\midrule
Alph.\ Soup + Tom.\ Sauce & 52.7\pms{5.8} & 14.0\pms{2.0} & \textbf{68.7\pms{5.0}} & \textbf{28.7\pms{8.1}} & 94.7\pms{5.8} & 68.7\pms{3.1} \\
Both Moka Pots $\to$ Stove & 49.3\pms{20.1} & 22.7\pms{8.3} & \textbf{70.0\pms{7.2}} & \textbf{75.3\pms{3.1}} & 84.0\pms{7.2} & 76.0\pms{7.2} \\
\midrule
\textbf{Average} & 51.0\pms{12.8} & 18.3\pms{4.0} & \textbf{69.3\pms{1.2}} & \textbf{52.0\pms{5.3}} & 89.3\pms{6.0} & 72.3\pms{2.5} \\
\bottomrule
\end{tabular}
\caption{Position perturbation success rate (\%) before and after curriculum training. Curriculum closes much of the gap with $\pibase^{50}$ without additional demonstrations. Bold marks improvement over non-curriculum $\pi_{\mathrm{RL}}^1$.}
\label{tab:curriculum_results}
\end{table}

These results confirm that progressively expanding the reset distribution during online RL can close much of the position-generalization gap without collecting additional demonstrations. Each curriculum stage builds on the behavioral repertoire acquired at the previous level, allowing the policy to incrementally extend its reach to larger displacements. In \cref{app:position_generalization_analysis}, we analyze what drives the position-generalization gap between $\pi_{\mathrm{RL}}^1$ and $\pibase^{50}$ and why the curriculum is effective.

\section{Position Generalization Analysis}
\label{app:position_generalization_analysis}
\begin{figure}[t]
    \centering
    \captionsetup{font=footnotesize}
    \captionsetup[subfigure]{font=footnotesize}
    \begin{subfigure}[t]{0.38\columnwidth}
        \centering
        \includegraphics[width=\linewidth]{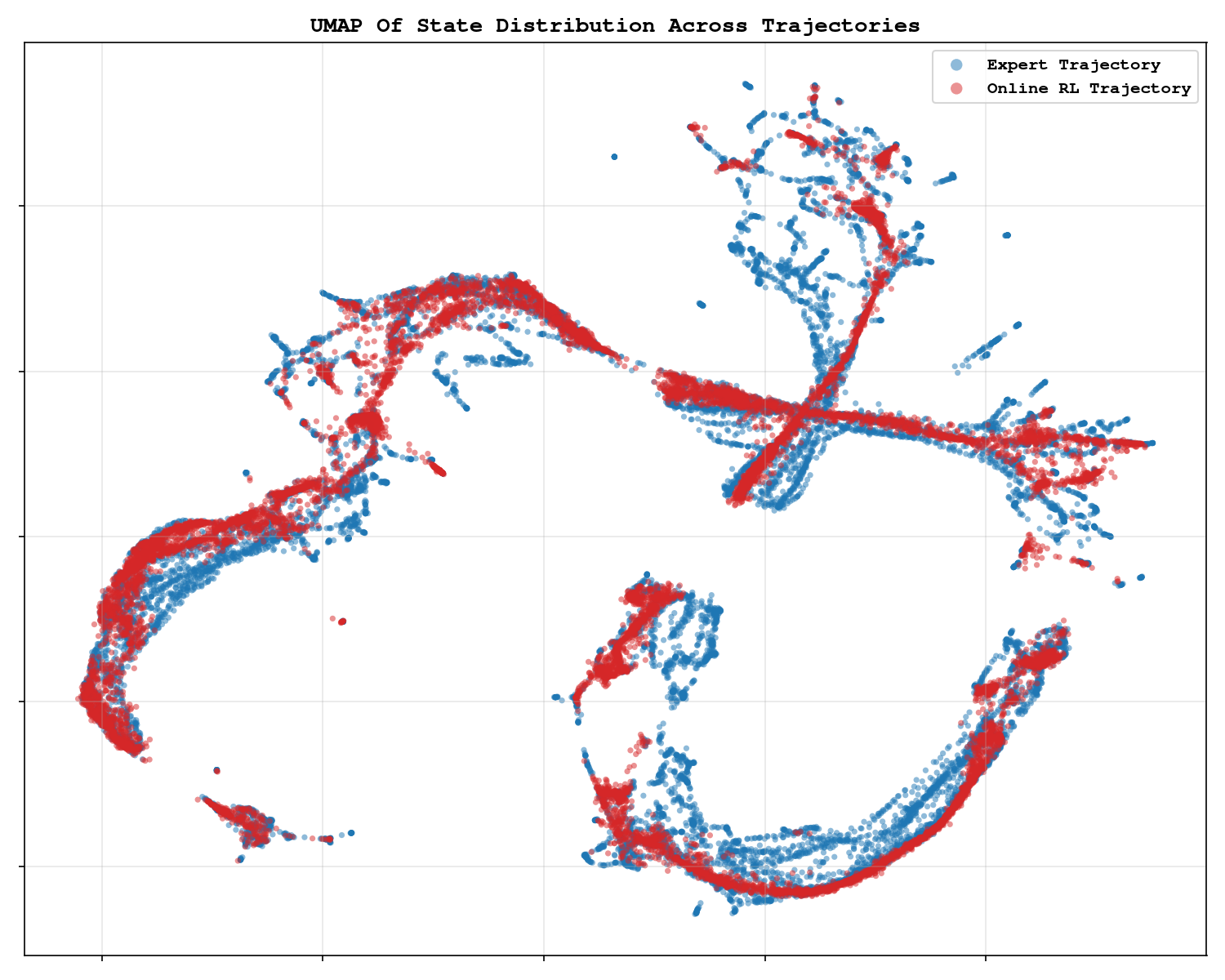}
        \caption{State-space coverage.}
        \label{fig:v2_umap_coverage}
    \end{subfigure}
    \hfill
    \begin{subfigure}[t]{0.38\columnwidth}
        \centering
        \includegraphics[width=\linewidth]{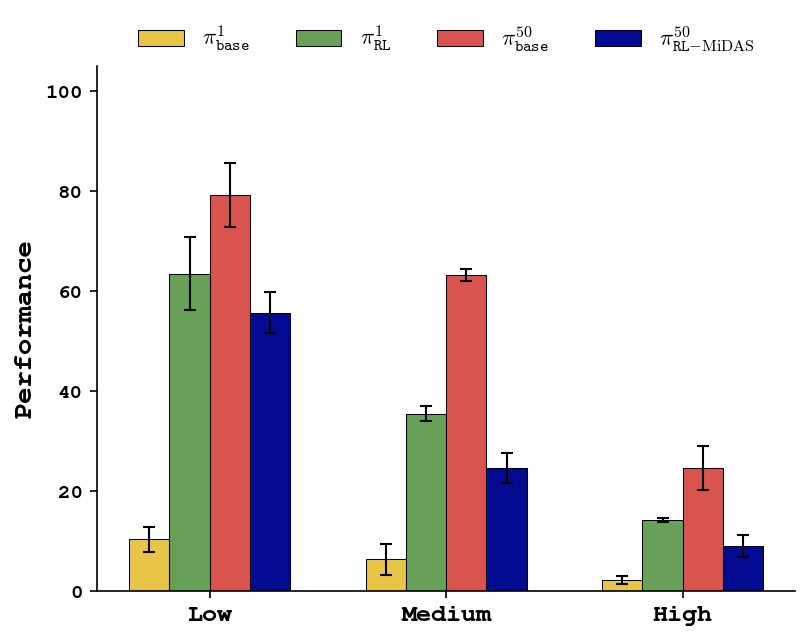}
        \caption{Position generalization.}
        \label{fig:v2_pos_gen}
    \end{subfigure}
    \caption{\textbf{(a)} $\pi_{rl}^1$ covers a narrow region of state space around its warm-start demo; $\pibase^{50}$ visits a much broader distribution. \textbf{(b)} Success degrades with displacement magnitude; distilling $\pi_{rl}^1$ rollouts into the base policy does not close the gap with $\pibase^{50}$.}
    \label{fig:v2_coverage_and_pos}
\end{figure}
\confpar{Position perturbations expose a gap between ID success and state-shift robustness.}
We isolate this effect using position perturbations at different magnitudes. We compare $\pi_{\mathrm{RL}}^1$ to $\pibase^1$ and to $\pibase^{50}$, a behavior-cloned policy trained on all 50 human-teleoperated demonstrations for the task. Although $\pi_{\mathrm{RL}}^1$ and $\pibase^{50}$ achieve similar in-distribution success, Fig.~\ref{fig:v2_pos_gen} shows that $\pibase^{50}$ generalizes much better as perturbations grow. \emph{The gap is not explained by the residual parameterization alone.} One possibility is that $\pi_{\mathrm{RL}}^1$ generalizes poorly because the residual actor is less expressive than the pretrained policy architecture. To test this, we distill 50 successful rollouts from $\pi_{\mathrm{RL}}^1$ back into the base policy, yielding $\pi_{\mathrm{RL-MIDAS}}^{50}$. This still does not recover the robustness of $\pibase^{50}$; in fact, $\pi_{\mathrm{RL-MIDAS}}^{50}$ performs worse than $\pi_{\mathrm{RL}}^1$ under position perturbations. This suggests that the gap is driven not by policy parameterization, but by differences in state coverage and motion priors between human expert demonstrations and autonomous data. We visualize this coverage in Fig.~\ref{fig:v2_umap_coverage}, where the 50 human demonstrations span a broader region of state space than successful autonomous trajectories, further supporting this explanation.

\section{Baselines}
\label{app:baselines}
\begin{table}[t]
    \centering
    \captionsetup{font=footnotesize}
    \caption{Success rate (\%) of flow-matching BC baselines on LIBERO-Long. Policies are evaluated at the final checkpoint on 50 canonical initial states.}
    \label{tab:libero-flow-bc-results}
    \begin{tabular}{lcc}
    \toprule
    Task & Image & State \\
    \midrule
    \multicolumn{3}{l}{\textit{LIBERO-Long}} \\
    \quad Alph.\ Soup + Cr.\ Cheese $\to$ Basket & 0.0 & 0.0 \\
    \quad Alph.\ Soup + Tom.\ Sauce $\to$ Basket & 0.0 & 0.0 \\
    \quad Black Bowl $\to$ Bottom Drawer & 14.0 & 0.0 \\
    \quad Book $\to$ Caddy & 0.0 & 0.0 \\
    \quad Both Moka Pots $\to$ Stove & 0.0 & 0.0 \\
    \quad Cr.\ Cheese + Butter $\to$ Basket & 0.0 & 0.0 \\
    \quad Moka Pot $\to$ Stove & 8.0 & 0.0 \\
    \quad White Mug + Choc.\ Pudding & 0.0 & 0.0 \\
    \quad White Mug + Plates & 0.0 & 0.0 \\
    \quad Yellow-White Mug $\to$ Microwave & 30.0 & 0.0 \\
    \midrule
    \textbf{Average} & \textbf{5.2} & \textbf{0.0} \\
    \bottomrule
    \end{tabular}
\end{table}
\subsection{Flow matching policy BC baseline}
We include two flow-matching behavior-cloning baselines on LIBERO to separate the effect of the pretrained VLA from the effect of using an expressive action decoder. Both baselines use the same flow-matching action generator and differ only in how observations are represented: one policy conditions on image observations and robot proprioception, while the other conditions on a privileged augmented state vector.

\paragraph{Flow-matching action generator.}
The action generator is a Multi-Modal Diffusion Transformer (MMDiT), following the joint-attention design used in Stable Diffusion 3-style generative models and adapted here from image generation to action-chunk prediction. We use this architecture as a design template: the flow policy itself is trained on LIBERO demonstrations, while the image-conditioned variant uses ImageNet-pretrained ResNet encoders for visual observations. The policy predicts a chunk of $H=16$ future actions, where each action is a 7-dimensional end-effector command consisting of $\Delta$position, $\Delta$orientation, and gripper control.

We train the policy with rectified-flow matching. Given an expert action chunk $a$ and Gaussian noise $z \sim \mathcal{N}(0,I)$, we sample a flow time $t \sim \mathcal{U}[0,1]$ and construct the linear interpolation
\[
x_t = t z + (1-t)a .
\]
The network is trained to regress the conditional velocity field along this path using an action-padding-masked MSE loss. At inference time, actions are sampled by integrating the learned velocity field from Gaussian noise using 10 Euler steps, and we use EMA weights for evaluation.

Observation tokens and noised action tokens are linearly embedded to a width of 768 and augmented with frozen 1D sinusoidal positional embeddings. The MMDiT backbone then applies joint attention over observation and action tokens, with the action tokens allowed to attend to all observation tokens. The backbone has 12 transformer blocks, 12 attention heads, head dimension 64, and an FFN expansion factor of 4. Each block uses timestep-conditioned adaLN-Zero modulation, joint multi-head self-attention with RMS-normalized queries and keys, and a feed-forward layer. A final modulated projection maps the action tokens back to $\mathbb{R}^7$.

\begin{table}[h]
    \centering
    \captionsetup{font=footnotesize}
    \scriptsize
    \caption{Shared hyperparameters for the flow-matching BC baselines.}
    \label{tab:flow-bc-hparams}
    \begin{tabular}{ll}
    \toprule
    Hyperparameter & Value \\
    \midrule
    Backbone & MMDiT \\
    Transformer depth & 12 \\
    Hidden size & 768 \\
    Attention heads & 12 \\
    Head dim. & 64 \\
    FFN expansion & 4 \\
    Obs. history & 2 \\
    Action horizon & 16 \\
    Action dim. & 7 \\
    Flow path & rectified flow \\
    Flow-time sampling & $\mathcal{U}[0,1]$ \\
    Inference steps & 10 Euler \\
    Optimizer & AdamW \\
    Learning rate & $10^{-4}$ \\
    Weight decay & $0.01$ \\
    EMA decay & $0.9999$ \\
    Normalization & z-score \\
    \bottomrule
    \end{tabular}
\end{table}

\paragraph{Image-conditioned flow baseline.}
The image-conditioned baseline uses two camera streams: a third-person \texttt{agentview} camera and a wrist-mounted \texttt{eye-in-hand} camera. Each camera is processed by an independent ImageNet-pretrained ResNet-18 encoder. We replace BatchNorm layers with GroupNorm and truncate the ResNet at \texttt{layer4}, producing a $512 \times 3 \times 3$ feature map per frame. Input images are resized to $100 \times 100$, ImageNet-normalized, and cropped to $84 \times 84$, using random crops during training and center crops during evaluation.

Each camera feature map is converted into a single observation token using a learned spatial-embedding head, followed by a linear bottleneck to the MMDiT width of 768. The policy also receives robot proprioception, consisting of end-effector position, end-effector axis-angle orientation, and gripper joint positions. This 8-dimensional proprioceptive vector is linearly embedded as one additional observation token. Thus, for each observation timestep the image policy receives three tokens: one token for each camera and one proprioceptive token. With a two-step observation history, the policy receives six observation tokens total.

\paragraph{State-conditioned flow baseline.}
The state-conditioned baseline only conditions on a privileged augmented state vector. This vector is used identically when reconstructing observations from dataset MuJoCo states and during online evaluation. The state vector is constructed from four blocks: robot proprioception, object/fixture/site poses, BDDL goal-predicate indicators, and goal-target geometry.

The proprioceptive block contains end-effector position, end-effector axis-angle orientation, and gripper joint positions, giving 8 dimensions. The entity-pose block contains the position and quaternion of every object, fixture, and site, sorted by name. The goal-predicate block contains one binary indicator per BDDL goal sub-predicate. The goal-target geometry block contains the world position of each goal sub-predicate's target entity. For a task with $N_{\mathrm{ent}}$ entities and $N_{\mathrm{goal}}$ goal sub-predicates, the resulting state dimension is
\[
D = 8 + 7N_{\mathrm{ent}} + 4N_{\mathrm{goal}} .
\]
The state-conditioned policy receives a two-step history of this full augmented state vector and embeds it into observation tokens for the same MMDiT action generator used by the image baseline. 

\begin{table}[h]
\centering
\captionsetup{font=footnotesize}
\scriptsize
\caption{Privileged-state dimensions for the LIBERO flow baseline.}
\label{tab:libero-state-dims}
\begin{tabular}{cccc}
\toprule
Task & $N_{\mathrm{ent}}$ & $N_{\mathrm{goal}}$ & $D$ \\
\midrule
0 & 17 & 2 & 135 \\
1 & 17 & 2 & 135 \\
2 & 7  & 2 & 65  \\
3 & 13 & 2 & 107 \\
4 & 10 & 2 & 86  \\
5 & 11 & 1 & 89  \\
6 & 10 & 2 & 86  \\
7 & 11 & 2 & 93  \\
8 & 7  & 3 & 69  \\
9 & 9  & 2 & 79  \\
\bottomrule
\end{tabular}
\end{table}

\subsection{Diffusion-Steering RL (DSRL)}

DSRL~\citep{wagenmaker2025steeringdiffusionpolicylatent} trains a SAC agent that operates in the noise space of the frozen $\pifive$ flow-matching policy. At each timestep, the SAC actor receives VLM features extracted by the frozen $\pifive$ backbone and outputs a perturbation to the initial noise vector; the frozen denoising head then converts the perturbed noise into an executed action chunk. Because the base policy weights remain fixed, all task-relevant adaptation occurs through the learned noise perturbation. We use the official codebase of \citet{wagenmaker2025steeringdiffusionpolicylatent} for this baseline.

The SAC actor and critic are three-layer MLPs with hidden dimension 128. Actions are predicted as chunks of length $C = 10$, and online updates begin after 500 initial environment steps. All other training infrastructure (evaluation protocol, reward definition, environment wrappers) matches the \midas Stage~II setup (\cref{app:impl_details}).

We note that for RoboCasa tasks evaluated at 150k steps, we estimate DSRL performance at the reporting point by linearly interpolating the evaluations at 140k and 160k steps separately
for each task and training seed. Each endpoint evaluation uses 50 rollouts. This is used for the numbers in \cref{tab:one_demo_results}.

\begin{table}[h]
\centering
\small
\caption{DSRL hyperparameters.}
\begin{tabular}{ll}
\toprule
\textbf{Hyperparameter} & \textbf{Value} \\
\midrule
Batch size & 64 \\
Discount factor ($\gamma$) & 0.999 \\
Actor learning rate & $10^{-4}$ \\
Critic learning rate & $3 \times 10^{-4}$ \\
Temperature learning rate & $3 \times 10^{-4}$ \\
Hidden dimensions & $(128, 128, 128)$ \\
Update-to-data ratio & 20 \\
Action chunk length ($C$) & 10 \\
Action noise magnitude & 1.0 \\
Online update start & 500 steps \\
Optimizer & Adam \\
\bottomrule
\end{tabular}
\end{table}

\subsection{Filtered BC}

Filtered BC iteratively collects autonomous rollouts from $\pibase^K$, retains only successful trajectories, and retrains the base policy on this filtered data. The model architecture and trainable parameters are identical to Stage~I behavior cloning (\cref{app:bc_details}).

\textbf{Procedure.}
Training proceeds in rounds. Each round consists of three phases: (1)~collect $N$ rollouts using the current policy, (2)~filter for trajectories that reached the task goal, and (3)~retrain the policy on the filtered set for $M$ gradient steps. The updated checkpoint from each round initializes the next. We use cumulative filtering, where successful trajectories accumulate across rounds so the training set grows monotonically. The $K$ expert demonstrations are included alongside the self-collected successes in every round. We run 20 rounds with 20 rollouts per round and 500 gradient steps per round, for a total of 400 collection rollouts and 10{,}000 gradient steps. We keep this setting as we find that the success rate saturates after the initial rise in our experiments and showed little further improvement within the evaluated budget.

\textbf{Trainable parameters.}
The same parameter groups are updated as in Stage~I: LoRA adapters (rank 32, $\alpha = 32$) on the PaliGemma 2B vision-language-model (VLM) backbone, full fine-tuning of the Gemma 300M action head, and full fine-tuning of the SigLIP vision encoder. All other backbone weights remain frozen.

\textbf{Optimization.}
Each round trains with the flow-matching loss using the optimizer and learning-rate schedule from the $\pifive$ training config (AdamW, $\beta_1 = 0.9$, $\beta_2 = 0.95$, peak learning rate $2.5 \times 10^{-5}$, cosine decay to $2.5 \times 10^{-6}$). The batch size is 16.

\begin{table}[h]
\centering
\small
\caption{Filtered BC hyperparameters.}
\begin{tabular}{ll}
\toprule
\textbf{Hyperparameter} & \textbf{Value} \\
\midrule
Number of rounds & 20 \\
Rollouts per round & 20 \\
Training steps per round & 500 \\
Total training steps & 10{,}000 \\
Batch size & 16 \\
Cumulative data & Yes \\
Include expert demos & Yes \\
Optimizer & AdamW \\
Peak learning rate & $2.5 \times 10^{-5}$ \\
LR schedule & Cosine decay \\
Minimum learning rate & $2.5 \times 10^{-6}$ \\
LoRA rank & 32 \\
LoRA $\alpha$ & 32 \\
LoRA targets & Attention + FFN \\
Action expert & Fully fine-tuned \\
Vision encoder (SigLIP) & Fully fine-tuned \\
Action horizon ($C$) & 10 \\
Loss & Flow matching \\
\bottomrule
\end{tabular}
\end{table}

\subsection{DICE-RL}

DICE-RL~\citep{sun2026priorproefficientskill} trains a residual actor using a TD3+BC-style objective on top of the frozen $\pifive$ base policy. The residual actor is parameterized as a tanh-Gaussian over flattened action chunks, and its output $\delta$ is added to the base action to form the executed action
\[
a_{\mathrm{exec}} = \mathrm{clip}\!\left(\abase + \alpha\,\delta,\, -1,\, 1\right),
\]
where $\alpha$ is a fixed residual scale. The base policy weights remain frozen; only the residual actor and a Q-critic are trained. All training infrastructure (reward definition, environment wrappers, evaluation protocol) matches the \midas Stage~II setup (\cref{app:impl_details}), and the policy conditions on the frozen $\pifive$ VLM embedding rather than raw pixels.

\textbf{Networks.}
The residual actor is an MLP tanh-Gaussian policy over the $C{\times}7$ action chunk ($C=10$), and the critic is an ensemble of $10$ state--action Q-heads. Both use hidden width $512$ and condition on the VLM embedding together with the base action. The critic is reduced by its ensemble mean, and a soft target critic is maintained by Polyak averaging ($\tau = 0.05$).

\textbf{Critic update.}
The critic regresses an entropy-free, $K$-sample bootstrap target. At each next state, $K = 16$ residuals are sampled from the actor, composed into executed actions, scored by the target critic, and averaged before the ensemble reduction, giving the target $r + \gamma\, \mathbb{E}_K[Q']$ with a sparse reward and an MSE loss. No entropy term is added to the backup.

\textbf{Actor update.}
The actor maximizes the mean of the critic over $K = 16$ action samples (a deterministic-policy-gradient-style objective with no entropy bonus), regularized by a \emph{filtered} zero-residual penalty $\lambda\,\alpha^2 \lVert \delta \rVert^2$. This anchor pulls the residual toward zero (i.e.\ toward the base action) except on transitions where the current action already improves on the base ($Q(s,a_{\mathrm{cur}}) \ge Q(s,\abase)$) and is calibrated to not be incorrectly overestimating the true Q value with ($Q(s,a_{\mathrm{cur}}) - \mathrm{MC} \le \epsilon$).

\textbf{Data and warmup.}
We run in demo-buffer mode: the success buffer is seeded from the single expert demonstration and frozen as an offline buffer, while the main buffer becomes online-only after a demo BC warmup phase (UTD $10$). Each minibatch mixes offline and online data, with the offline fraction decayed linearly from $0.5$ to $0.1$ over the first $500$ gradient steps. Online updates begin after $500$ chunked environment steps, and each gradient step applies $20$ critic updates and $10$ actor updates per chunked environment step.

\begin{table}[h]
\centering
\small
\caption{DICE-RL hyperparameters.}
\begin{tabular}{ll}
\toprule
\textbf{Hyperparameter} & \textbf{Value} \\
\midrule
Batch size & 64 \\
Discount factor ($\gamma$) & 0.999 \\
Actor learning rate & $10^{-4}$ \\
Critic learning rate & $3 \times 10^{-4}$ \\
Hidden dimensions & 512 \\
Critic ensemble size & 10 \\
Critic reduction & mean \\
Target update rate ($\tau$) & 0.05 \\
Optimizer & Adam \\
Max grad norm & 1.0 \\
Action chunk length ($C$) & 10 \\
Residual scale ($\alpha$) & 0.5 \\
Gradient steps per env step (UTD) & 1 \\
Critic updates per step & 20 \\
Actor updates per step & 10 \\
$K$ action samples & 16 \\
Zero-residual penalty coeff.\ ($\lambda$) & 1.0 \\
Filter threshold ($\epsilon$) & $-0.5$ \\
Online update start & 500 steps \\
Offline buffer ratio (start $\to$ end) & $0.5 \to 0.1$ \\
Ratio decay horizon & 500 steps \\
Success buffer min.\ size & 100 \\
Demo BC warmup UTD & 10 \\
Number of demos & 1 \\
Conditioning & $\pifive$ VLM embedding \\
\bottomrule
\end{tabular}
\end{table}

\section{Shared Principles Between \midas and Dice-RL}
\label{app:dicerl_shared_principles}

MiDAS and concurrent DICE-RL~\citep{sun2026priorproefficientskill} share several underlying principles for stable online adaptation of pretrained policies. Both retain a behavior-cloned policy and its observation encoder as frozen components, and learn lightweight action-space corrections using chunk-level critics. This reuses previously learned behavioral and perceptual structure while concentrating online learning on corrective control. Both also balance successful demonstration data with newly collected experience. MiDAS oversamples a success buffer initialized with demonstrations and expanded with successful online rollouts; DICE-RL mixes offline demonstrations with online replay and gradually reduces the demonstration fraction during an initial training period. These distinct replay strategies reflect a shared principle of retaining useful demonstration data while learning from autonomous experience, which is particularly relevant in our setting where initial success is rare. Finally, both methods balance behavioral anchoring with the freedom to learn corrections: MiDAS initializes its actor to reproduce base proposals before applying PA-RL updates, while DICE-RL selectively relaxes its residual regularization when value estimates and a Monte Carlo consistency check support improvement over the base action. These commonalities connect the two methods at the level of frozen representations, value-guided action correction, and careful use of demonstration data. Their strong performance in our one-demonstration experiments provides complementary evidence for the effectiveness of this combination in minimal-data adaptation.

\section{Real World Implementation Details}
\label{app:real_world_exps}
For our real world experiments, we use a bimanual YAM platform and evaluate \midas on two tasks: \textbf{(T1)} placing a green block in the right container and a blue block in the left container, and \textbf{(T2)} placing a knife and a donut on a plate. We fine-tune the $\pifive$ base model for 20{,}000 gradient steps using LoRA adapters for both the vision-language-model (VLM) backbone and the action head. The key differences from the simulation setup (\cref{app:bc_details}) are described below.

\textbf{Observation space.}
The robot receives RGB images from three cameras: a top-down view and two wrist-mounted cameras (left and right). The state observation consists of the joint positions and gripper states of both arms.

\textbf{Action space.}
Actions are predicted as chunks of horizon $C{=}60$, longer than the simulation horizon of 10 to accommodate the lower control frequency and extended task durations on hardware. Joint position targets use delta actions while gripper commands are absolute, matching the bimanual action space structure (6 joints + 1 gripper per arm).

\textbf{Trainable parameters.}
Unlike the simulation configuration where the Gemma 300M action head is fully fine-tuned (\cref{app:bc_details}), the real-world setup applies LoRA adapters to both the PaliGemma 2B vision-language-model (VLM) backbone and the Gemma 300M action head. This reduces the number of trainable parameters and limits overfitting to the single available demonstration per task.

\textbf{Optimization.}
The learning rate follows a cosine decay schedule with 1{,}000 warmup steps, a peak learning rate of $2.5 \times 10^{-5}$, and a minimum learning rate of $2.5 \times 10^{-6}$ reached after 10{,}000 decay steps. EMA averaging is disabled. \Cref{tab:real_world_hyperparams} lists all hyperparameters.

\textbf{Training time and rollouts collected.} We train our models with \midas for 5-6 hours, during the course of which we collect 350 trajectories for each task.

\begin{table}[h]
\centering
\small
\caption{Real-world behavior cloning hyperparameters for Stage~I finetuning on the YAM platform.}
\label{tab:real_world_hyperparams}
\begin{tabular}{ll}
\toprule
\textbf{Hyperparameter} & \textbf{Value} \\
\midrule
Number of demonstrations ($K$) & 1 \\
Training steps & 20{,}000 \\
Peak learning rate & $2.5 \times 10^{-5}$ \\
LR schedule & Cosine decay \\
LR warmup steps & 1{,}000 \\
Minimum learning rate & $2.5 \times 10^{-6}$ \\
LR decay steps & 10{,}000 \\
EMA decay & None \\
PaliGemma 2B backbone & LoRA \\
Action expert (Gemma 300M) & LoRA \\
Action horizon ($C$) & 60 \\
Action type & Delta (joints), absolute (grippers) \\
Cameras & Top, left wrist, right wrist \\
Checkpoint save interval & 5{,}000 \\
\bottomrule
\end{tabular}
\end{table}

\textbf{Stage II: Online residual RL.}
For Stage~II, the residual actor-critic is trained on top of the frozen BC-fine-tuned policy using PARL (\cref{app:parl_details}). The architecture follows the simulation setup with adjustments for the bimanual action space ($d_a{=}14$, comprising 6 joints and 1 gripper per arm) and the longer action horizon. The actor and critic are 4-layer MLPs with 1024 hidden units per layer, wider than the 3-layer 512-unit networks used in simulation. The residual policy is queried every 30 steps within each 60-step base action chunk after which is new chunk is sampled.

Several PARL hyperparameters are scaled relative to the simulation configuration (\cref{tab:parl_hyperparams}). The PA-RL procedure samples 32 candidates and retains 10 elites, and the critic and actor receive 40 and 20 updates per chunked environment step, respectively. A per-joint trust-region constraint of 0.2\,rad limits the distance between refined actions and the base policy proposal for non-gripper dimensions; grippers are uncapped. \Cref{tab:real_world_parl_hyperparams} lists all Stage~II hyperparameters.

\begin{table}[h]
\centering
\small
\caption{Real-world PARL hyperparameters for Stage~II online RL on the YAM platform. Entries that differ from the simulation defaults (\cref{tab:parl_hyperparams}) are marked with $^\dagger$.}
\label{tab:real_world_parl_hyperparams}
\begin{tabular}{ll}
\toprule
\textbf{Hyperparameter} & \textbf{Value} \\
\midrule
Batch size ($B$)$^\dagger$ & 128 \\
Discount factor ($\gamma$)$^\dagger$ & 0.9999 \\
Target network update rate ($\tau$) & 0.05 \\
Actor learning rate ($\alpha_\pi$) & $10^{-4}$ \\
Critic learning rate & $3 \times 10^{-4}$ \\
Hidden dimensions$^\dagger$ & $(1024, 1024, 1024, 1024)$ \\
VLM embedding bottleneck dim & 200 \\
Number of Q-networks ($J$) & 10 \\
Critic ensemble reduction & mean \\
PA-RL candidate samples ($N$)$^\dagger$ & 32 \\
PA-RL elite samples ($M$)$^\dagger$ & 10 \\
PA-RL gradient steps ($G$) & 30 \\
PA-RL step size ($\eta$) & $10^{-3}$ \\
PA-RL trust-region radius$^\dagger$ & 0.2\,rad (joints), uncapped (grippers) \\
Critic updates per chunked environment step ($U_C$)$^\dagger$ & 40 \\
Actor updates per chunked environment step ($U_A$)$^\dagger$ & 20 \\
Warmup chunked environment steps & 10{,}000 \\
\shortstack[l]{BC warmup critic/actor updates per\\chunked environment step ($U_C^{\mathrm{w}}, U_A^{\mathrm{w}}$)} & 8 \\
Success buffer ratio ($\rho_{\mathrm{succ}}$)$^\dagger$ & 0.20 \\
Success buffer minimum size$^\dagger$ & 100 \\
Gradient norm clip & 1.0 \\
Optimizer & Adam \\
Action chunk length ($C$, $H$) & 60 \\
Residual query frequency & 30 \\
Action dimension ($d_a$)$^\dagger$ & 14 \\
\bottomrule
\end{tabular}
\end{table}

\textbf{Details on evaluation and resets.}: For both online RL training and evaluations, the target objects are reset around the demonstration's target locations. We do two classes of evaluations: in-distribution, where the target objects are perturbed very mildly from their demonstration positions/orientations; out-of-distribution with larger perturbations. Videos of rollouts and training is available here: \href{https://minimal-data-adaptation.github.io/}{minimal-data-adaptation.github.io}.

\section{Limitations}

The results in this paper should be interpreted within two limitations. First, the present instantiation remains limited on longer-horizon tasks, where sparse-reward value learning must assign credit through long sequences of actions proposed by a frozen prior. Scaling MDA may therefore require stronger value priors, such as pretrained critics or value decompositions that expose intermediate progress signals.
Second, Stage II remains reset-dependent and interaction-intensive. This is manageable in simulation, but becomes a central constraint on hardware, where failures, resets, and safety limit autonomous practice without an operator in the loop. Reducing this dependence will require safer exploration, better failure detection, learned resets, or lightweight intervention mechanisms that preserve the low-demonstration setting. We discuss directions for future work in \cref{app:future_work}.

\end{document}